\documentclass{article}

\usepackage[margin=1in]{geometry}
\usepackage{mathpazo}

\usepackage[utf8]{inputenc} %
\usepackage[T1]{fontenc}    %
\usepackage{hyperref}       %
\usepackage{url}            %
\usepackage{booktabs}       %
\usepackage{amsfonts}       %
\usepackage{nicefrac}       %
\usepackage{microtype}      %
\usepackage{xcolor}         %
\usepackage{graphicx}
\usepackage[square,numbers]{natbib}
\usepackage{amsmath}
\usepackage{amssymb}
\usepackage{cleveref}
\usepackage{multicol}
\usepackage{comment}

\newcommand{\Duniv}{\widehat{\mathcal{D}}_{univ}}
\newcommand{\Dadv}{\widehat{\mathcal{D}}_{adv}}
\newcommand{\Drand}{\widehat{\mathcal{D}}_{rand}}
\newcommand{\Ddet}{\widehat{\mathcal{D}}_{det}}
\newcommand{\uap}{UAP}
\newcommand{\uaps}{UAPs}
\newcommand{\ltwo}{\ell_2}
\newcommand{\linf}{\ell_{\infty}}

\makeatletter
\def\@fnsymbol#1{\ensuremath{\ifcase#1\or \dagger\or \ddagger\or
   \mathsection\or \mathparagraph\or \|\or **\or \dagger\dagger
   \or \ddagger\ddagger \else\@ctrerr\fi}}
    \makeatother

\title{On Distinctive Properties of Universal Perturbations}

\author{%
  Sung Min Park\\
  MIT\\
  \texttt{sp765@mit.edu}\\
  \and
  Kuo-An Wei\thanks{Work done while author was a student at MIT.}\\
  MIT\\
  \texttt{kuoanwei@mit.edu}\\
  \and
  Kai Xiao\\
  MIT\\
  \texttt{kaix@mit.edu}\\
  \and
  Jerry Li\\
  Microsoft Research\\
  \texttt{jerrl@microsoft.com}\\
  \and
  Aleksander M\k{a}dry\\
  MIT\\
  \texttt{madry@mit.edu}\\
}

\date{}

\begin{document}

\maketitle

    \begin{abstract}
      
We identify properties of universal adversarial perturbations (UAPs) that distinguish them from standard adversarial perturbations.
Specifically, we show that targeted UAPs generated by projected gradient descent exhibit two human-aligned properties: semantic locality and spatial invariance, which standard targeted adversarial perturbations lack.
We also demonstrate that UAPs contain significantly less signal for generalization than standard adversarial perturbations---that is, UAPs leverage non-robust features to a smaller extent than standard adversarial perturbations.

    \end{abstract}

    \section{Introduction}
    \label{sec:intro}
    Modern deep neural networks perform extremely well across many prediction tasks, but they can be vulnerable to adversarial examples \citep{szegedy2014intriguing}.
This lack of robustness to such small, imperceptible perturbations shows that these models do not necessarily make predictions in a human-aligned way \citep{ilyas2019adversarial}.
A large body of work has studied various aspects of adversarial examples, from attacks \citep{papernot2016transferability, carlini2017towards} and defenses \citep{madry2017towards, lecuyer2019certified} to their possible origins \citep{goodfellow2014explaining, fawzi2016robustness, schmidt2018adversarially,
ilyas2019adversarial}.

An important subclass of these adversarial examples are \emph{universal adversarial perturbations} (\uaps{}) \citep{moosavi2017universal}, which are adversarial perturbations that are effective on a large fraction of inputs. Such \uaps{} can be used to execute an effective attack as they can be computed in advance, independently of the target input. This has inspired work on different approaches to efficiently generating UAPs \citep{mopuri2017fast, liu2019universal, khrulkov2018art, poursaeed2018generative, hayes2019learning, wu2019universal}.

\subsection{Our results}

In this paper, we study \uaps{} from a different perspective: understanding how they differ from standard adversarial perturbations. Specifically, we focus on \emph{targeted} perturbations generated by projected gradient descent (PGD), for both \uaps{} and standard adversarial perturbations.

\begin{figure}[!htbp]
  \centering
  \includegraphics[width=0.5\linewidth]{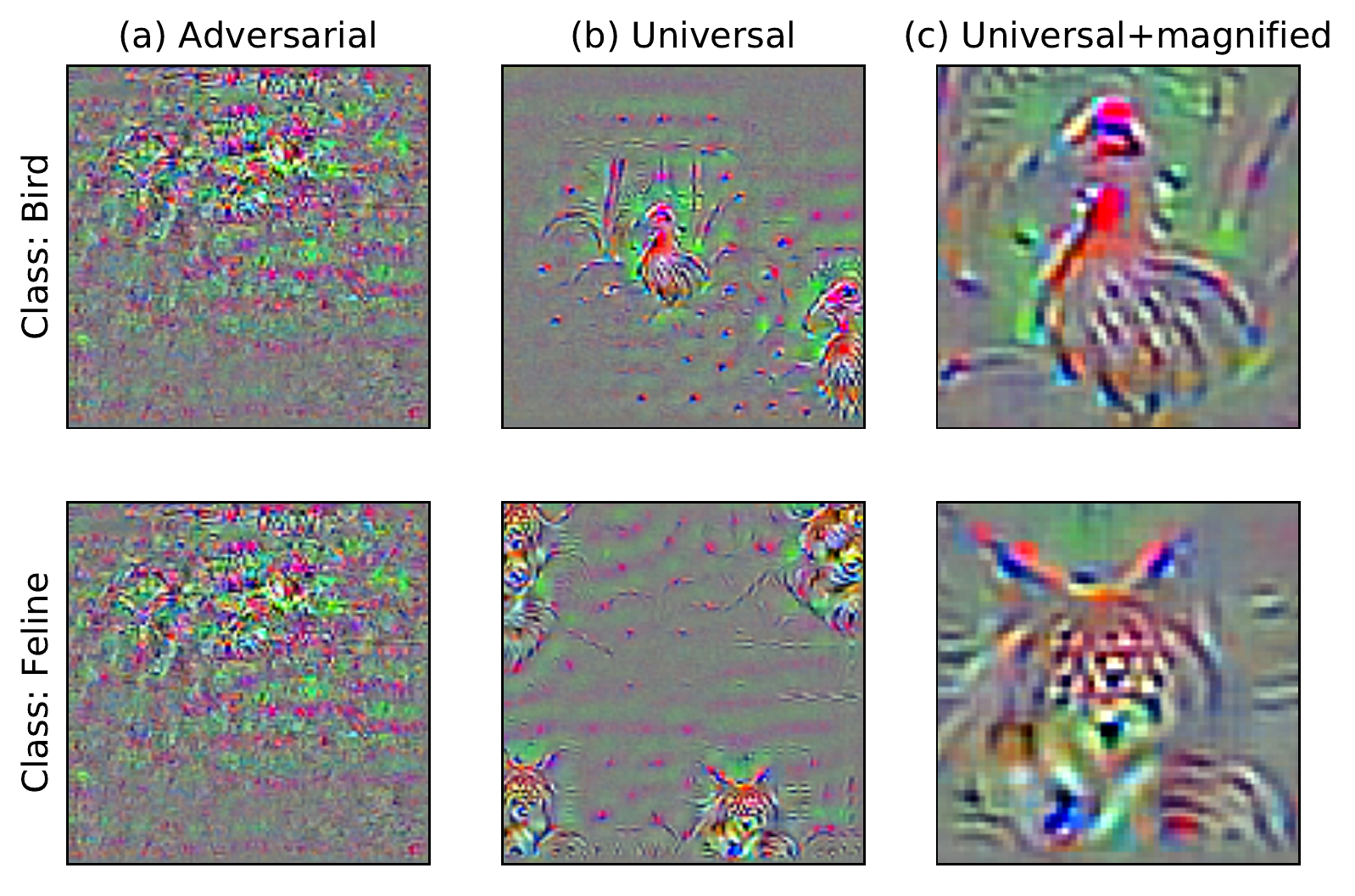}
  \caption{(Magnified) $\ell_2$ adversarial perturbations ($\epsilon=6.0$)
  towards two ImageNet-M10 classes, bird and feline: (a)
  \textbf{standard} adversarial perturbations towards two different classes generated on the same image, (b) \textbf{universal} adversarial perturbations, and (c)
  zooming in on the most semantic patch of (b).}
  \label{fig:adv_univ}
\end{figure}

We first observe that---in contrast to standard adversarial perturbations that tend to be incomprehensible---\uaps{} are more human-aligned (\Cref{fig:adv_univ}).\footnote{Prior works \citep{hendrik2017universal, hayes2019learning, khrulkov2018art, liu2019universal,zhang2020understanding} have made similar observations.}
We dissect this phenomenon further by demonstrating that \uaps{} have two distinctive properties:
(1) they are \emph{locally semantic}, in that the
signal is concentrated in local regions that are most salient to humans; and
(2) they are approximately \emph{spatially invariant}, in that they are still effective after translations.
These properties make \uaps{} better aligned with human priors and distinguish them from standard perturbations.

Next, we study the extent to which \uaps{} leverage {\em non-robust features} \citep{ilyas2019adversarial}, which are exploited by standard adversarial perturbations \citep{ilyas2019adversarial}.
These are features that are sensitive to small perturbations, yet still correlated with the correct label \citep{tsipras2019robustness}---in aggregate, their signal can be sufficient to generalize well on the underlying classification task.

As \uaps{} are better aligned with human priors, we might expect \uaps{} to contain signal that is more useful for generalization than standard adversarial perturbations. %
However, we find that this is not the case: \uaps{} contain significantly less generalizable signal from non-robust features compared to standard perturbations.
We quantify this by following the methodology of \citet{ilyas2019adversarial} and measuring
 (1) how well a model can generalize to the original test set by training on a dataset where the only correlations with the label are added via \uaps{}; and
 (2) measuring the transferability of \uaps{} across independently trained models.
Under these metrics, \uaps{} consistently obtain non-trivial but substantially worse generalization performance than standard adversarial perturbations.

    \section{Preliminaries}
    \label{sec:prelim}
    We consider a standard classification task: given input-label samples $(x,y) \in
\mathcal{X} \times \mathcal{Y}$ from a data distribution $\mathcal{D}$, the goal is to
learn a classifier $C : \mathcal{X} \rightarrow \mathcal{Y}$ that generalizes to new data.

\subsection{Definitions}
We focus on targeted perturbations generated by projected gradient descent (PGD).

\paragraph{Universal Adversarial Perturbations.}
A \emph{universal adversarial perturbation} \citep{moosavi2017universal}, or \emph{UAP}, is a
perturbation $\delta \in \Delta$ that causes the classifier $C$ to predict the wrong label on a large fraction of inputs from $\mathcal{D}$, where $\Delta$ is the set of allowed perturbations.
In contrast, a {\em standard adversarial perturbation} causes $C$ to predict the wrong label on one
specific input.
The only technical difference between \uaps{} and standard adversarial
perturbations is the use of a single perturbation
vector $\delta$ that is applied to all inputs. Thus, one can
think of universality as a constraint for the perturbation $\delta$
to be input-independent.

\paragraph{Targeted perturbations.}
We focus on \emph{targeted} \uaps{} that fool $C$ into predicting a
specific (usually incorrect) target label $t$. Thus, a (targeted) \uap{} $\delta \in \Delta$ satisfies $\mathbb{P}_{(x, y) \sim \mathcal{D}}[C(x+\delta) = t] = \rho$,
where $\rho$ is the \emph{attack success rate}\footnote{Technically, we only compute a finite sample approximation of ASR on the test set.} (ASR) of the \uap{}.\footnote{There is no natural cutoff for $\rho$ to make a certain perturbation universal vs. not; this depends on context.}
Targeted \uaps{} are the natural choice for our study here, as they allow us to isolate features (positively) correlated with a particular target.
For a fair comparison, we also only consider targeted versions of standard adversarial perturbations.
While we do not study untargeted \uaps{}, they generally perturb most examples towards a
single common target  \citep{moosavi2017universal}; this suggests that untargeted \uaps{} may have
similar properties to the targeted versions.

\paragraph{$\ell_p$ perturbations.}
We study the case where $\Delta$ is the set of $\ell_p$-bounded perturbations, i.e.
\[\Delta = \{\delta \in \mathbb{R}^d\:|\: ||\delta||_p \leq \varepsilon\}\] for $p=2,\infty$ where $\varepsilon > 0$.
This is the most widely studied setting for research on adversarial examples and has proven
to be an effective benchmark \citep{carlini2019on}.
Additionally, $\ell_p$-robustness appears
to be aligned to a certain degree with the human visual system \citep{tsipras2019robustness}.

\paragraph{Projected gradient descent.}
We focus on the family of perturbations computed by projected gradient descent (PGD), a standard method for generating adversarial examples.
While there exists other methods for computing \uaps{}, which may have different properties, focusing on this family allows us to isolate
universality as the sole factor of variation.
From here on, we refer to our specific class of targeted, PGD-generated \uaps{}
simply as \uaps{}.

\subsection{Computing \uaps{}}
\label{sec:prelim_computing}
We compute \uaps{} by using mini-batch projected gradient descent (PGD) to solve the following optimization problem:
\begin{equation}
\min_{\delta \in \Delta} \mathbb{E}_{(x, y)\sim\mathcal{D}}
\Big[ \mathcal{L}(f(x+\delta), t) \Big] \label{eq:loss}
\end{equation}
where $\mathcal{L}$ is the standard cross-entropy loss for classification, $f$ is the logits prior to classification,
and $t$ is the target label. While many different algorithms have been developed for computing \uaps{} with varying ASRs,
we use this simple algorithm instead as achieving the highest ASR is orthogonal to our investigations.
As observed in \citet{moosavi2017universal}, \uaps{} trained on only a fraction of the dataset still generalize to the rest of the dataset, so in practice it suffices to approximate the expectation in \eqref{eq:loss} using a relatively small batch of samples drawn from $\mathcal{D}$. We call the batch we optimize over the \emph{base set}, and use $K$ to refer to its size.
We carefully choose hyperparameters (\Cref{app:hyper}) to ensure that our results remain robust.%

\subsection{Experimental Setup}
\paragraph{Datasets.}
We conduct our experiments on a subset of ImageNet \citep{imagenet_cvpr09} and CIFAR-10 \citep{krizhevsky2009learning}.
The ImageNet Mixed10 dataset \citep{robustness}, which we refer to as ImageNet-M10, is formed by sub-selecting and grouping together semantically similar classes from ImageNet;
it provides a more computationally efficient alternative to the
full dataset, while retaining some of ImageNet's complexity (see~\Cref{app:dataset} for more details).
In the main text, we focus on ImageNet-M10, and in~\Cref{app:vis} we provide selected results on CIFAR-10 and full ImageNet.

\paragraph{Models.} We mainly focus on the standard ResNet-18 architecture \citep{he2016deep}, and highlight results obtained with other architectures in \Cref{app:other_arch}.

    \section{Quantifying Human-Alignment}
    \label{sec:properties}
\begin{figure*}[!htbp]
  \centering
  \includegraphics[width=0.833\linewidth]{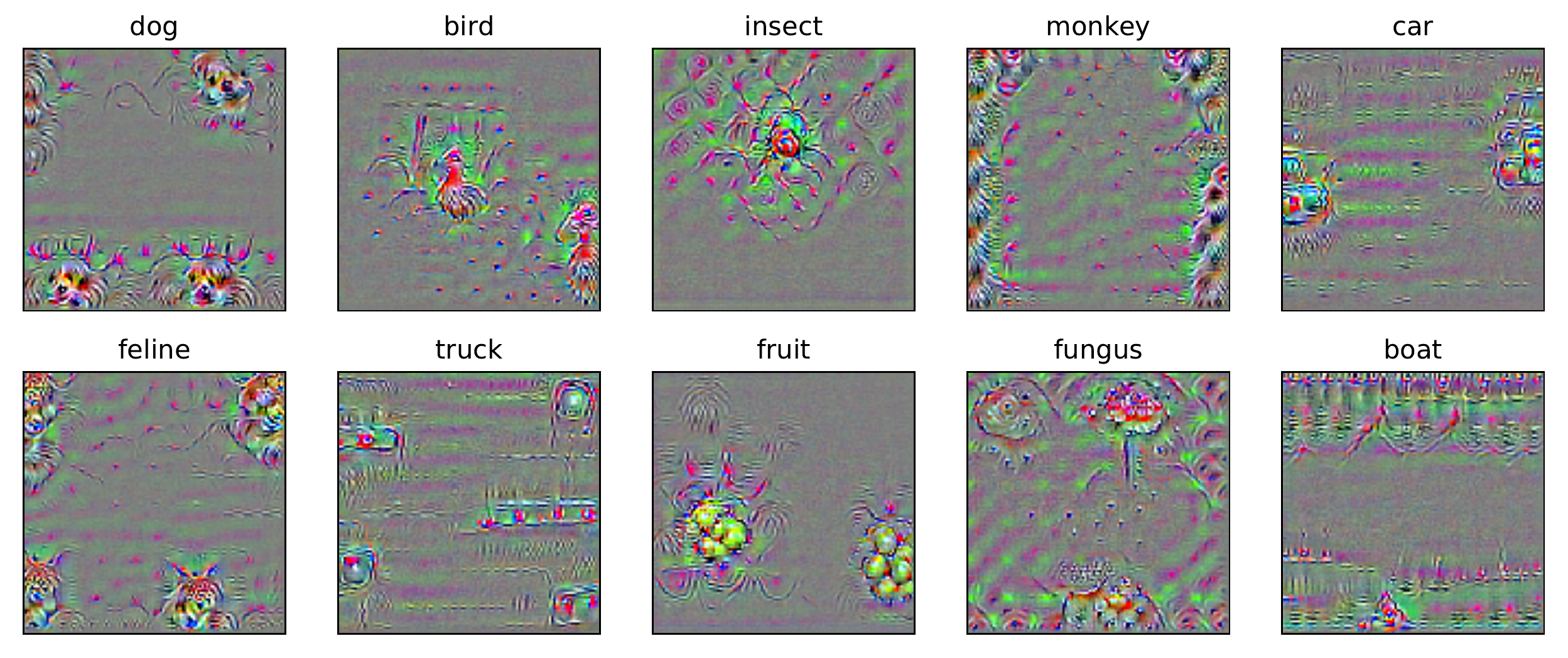}
  \caption{A sample of $\ell_2$ \uaps{} ($\epsilon=6.0$)
  for all ten ImageNet-M10 classes. We observe patterns resembling dog and cat (feline) heads, birds, outlines of cars and trucks, an insect, and mushrooms (fungus).}
  \label{fig:univ}
\end{figure*}

    Standard $\ell_p$ adversarial perturbations are often thought to be
incomprehensible to humans \citep{szegedy2014intriguing,ilyas2019adversarial}. Even when magnified for visualization, these
perturbations are not identifiable to a human as belonging to their target class.  In
contrast, \uaps{} are visually much more interpretable:
when amplified, they contain local regions that we can identify with the target class (see~\Cref{fig:univ} for example
\uaps{}, and Figure~\ref{fig:adv_univ} for a comparison to
standard perturbations). Prior works have similarly observed that \uaps{}
resemble their target class \citep{hayes2019learning} or segmentation \citep{hendrik2017universal} and contain meaningful texture patterns \citep{khrulkov2018art, liu2019universal}

We quantify the extent to which \uaps{} are human-aligned beyond their visualizations.
We identify two additional properties of \uaps{}---(1) {\em semantic locality} and (2) {\em spatial invariance}---which
are well-aligned with human vision priors.

\subsection{Semantic locality}
We first find that \uaps{} {\em locally semantic}, in that a significant fraction of the signal in the perturbation is concentrated in small, localized regions that are salient to humans.
This makes the perturbations more interpretable since the human visual system is known to attend to localized regions of images \citep{rensink2000dynamic}; indeed, this property also motivates some attribution methods in interpretability \citep{olah2018the}.
Hence, we aim to quantify the extent to which \uaps{} possess this property.

While this property might seem to follow from the earlier observation that \uaps{} contain local regions that are semantically identifiable as the target class and standard perturbations do not (cf. \Cref{fig:univ}), this is not so obvious;
a priori, it is unclear which parts of the perturbation influence the model.
Here, we find that the \uaps{} are indeed locally semantic, in that most of their signal comes from the most visually meaningful regions.
In contrast, standard perturbations lack this property as no local regions are semantic (as far as our current understanding of them).

\begin{figure*}[]
  \centering
  \includegraphics[width=\linewidth]{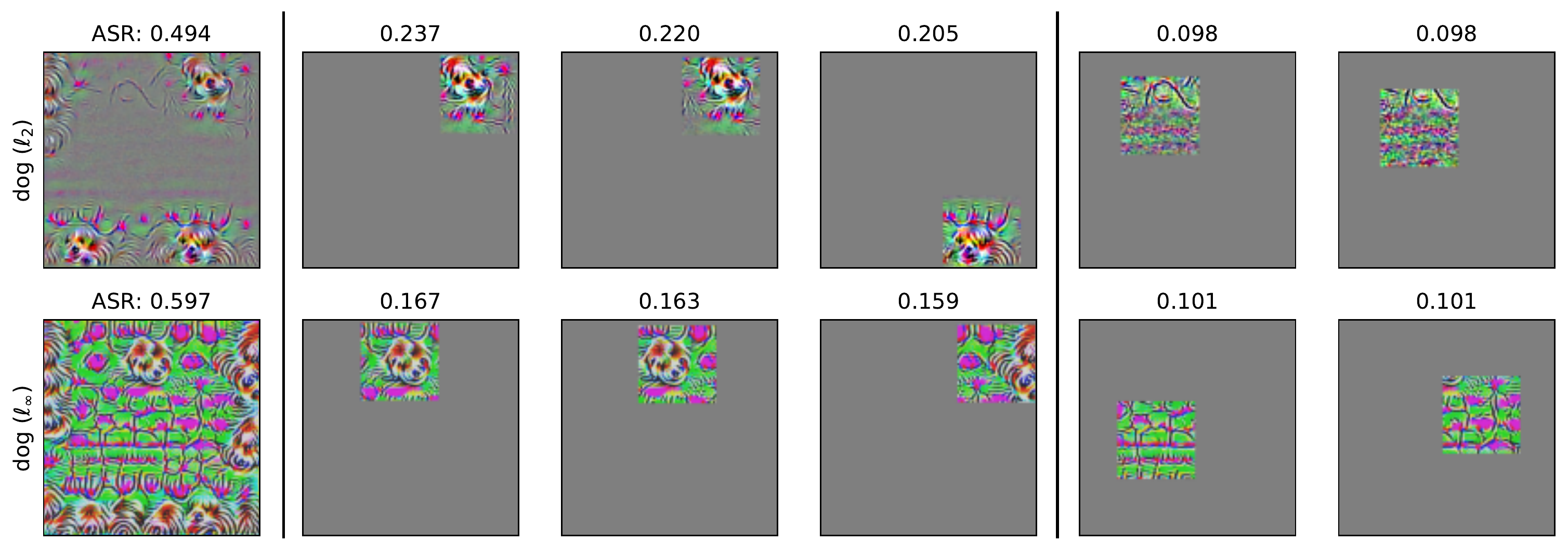}
  \caption{A randomized, local analysis of \uaps{} on ImageNet-M10. 64 random 80$\times$80 patches are isolated and evaluated on their ASR. First column shows the original perturbation for the target class (\textbf{top}: $\ltwo$; \textbf{bottom}: $\linf$); next five columns show three patches with the highest ASR, and two with the lowest. The number on top of each perturbation indicates its ASR on the test set. Patches are normalized to have the same norm as the highest norm patch.}
  \label{fig:manual}
\end{figure*}

\paragraph{Our methodology.}
To quantify this for UAPs,
we \emph{randomly} select local patches of the perturbation, evaluate their attack success rate (ASR) in isolation, and inspect them visually. For both $\ltwo$ and $\linf$ perturbations, the patches with the highest ASR are more visually identifiable as the target class
(Figure~\ref{fig:manual}).\footnote{For $\ltwo$ perturbations, the patches vary widely in norm, so a possible concern is that the model is only reacting to the patches with the highest norm. To account for this,
we linearly scale up all patches to have the same norm as the largest norm patch. For $\linf$ perturbations, all patches have similar norms, so no further normalization is done.}
This shows that the model is indeed influenced primarily by the most salient parts of the perturbation.

\subsection{Spatial invariance}
\label{sec:spatial}
We find that \uaps{} have the property of being {\em spatially invariant} to a large degree.
As spatial invariance is one of the key
properties of the human visual system \citep{hubel1968receptive},
it is also a desirable, human-aligned property.
In contrast, we demonstrate that standard adversarial perturbations are much more brittle to translations.
Hence, this illustrates another way that \uaps{} are more human-aligned than standard perturbations.
\paragraph{Our methodology.}
We quantify spatial invariance by measuring the ASR of
translated perturbations. A highly spatially invariant perturbation
will have a high ASR even after translations.
Specifically, we compute targeted
standard adversarial perturbations on a sample of 256 images from the ImageNet-M10 test
set. Then, we evaluate the ASR of different translated copies of
these perturbations over the respective images (e.g. translated copies of each perturbation are evaluated only on its corresponding image).
For comparison, we take a precomputed set of \uaps{} of the same norm, and evaluate them across all 256 images.
Translated perturbations use wrap-around to preserve information.\footnote{We also evaluated without wrap-around, and the results are similar.}

\begin{figure*}[]
  \centering
  \includegraphics[width=\linewidth]{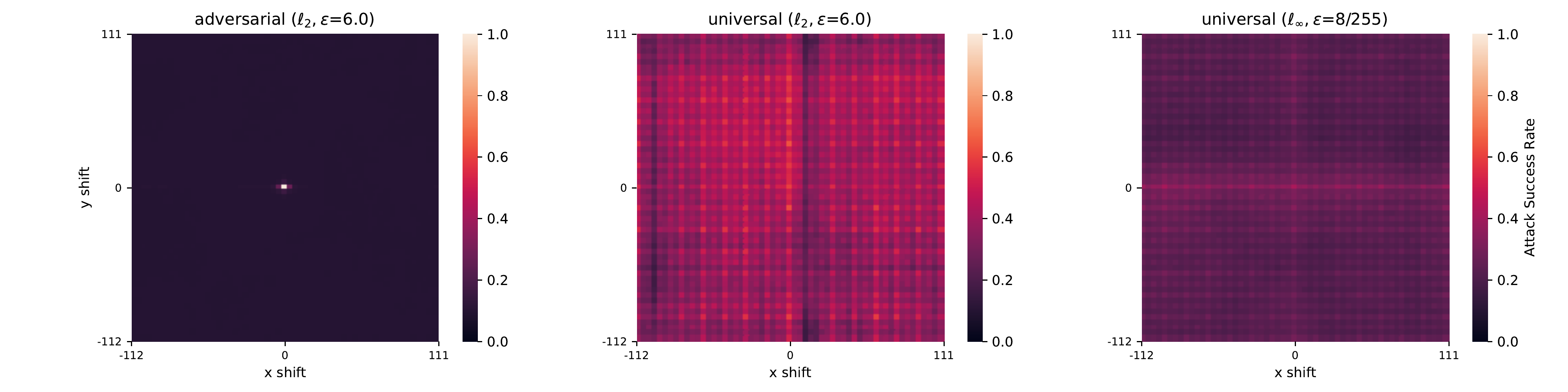}
  \caption{
    Evaluation of translated adversarial and
    universal $\ltwo, \linf$ perturbations for the ImageNet-M10 class \texttt{bird}.
    We evaluate a subsampled grid with strides of four pixels.
    The value at coordinate $(i, j)$ represents the average ASR when the
    perturbations are shifted right by $i$ pixels and up by $j$, with wrap-around to preserve information; the center pixel at $(0, 0)$ represents the ASR of the original unshifted perturbations. For \uaps{}, the ASR at each location is averaged over ten perturbations.
  }
  \label{fig:spatial}
\end{figure*}

Figure~\ref{fig:spatial} shows that \uaps{} still achieve
non-trivial ASR after translations of varying magnitudes. In contrast, standard adversarial perturbations achieve a chance-level $10\%$
ASR when shifted by more than eight pixels (two grid cells
in Figure~\ref{fig:spatial}). This illustrates that \uaps{}
are somewhat robust to translations, and thus are more spatially invariant
when compared to standard adversarial perturbations.
In~\Cref{app:symmetry}, we discuss why we might expect to see such an invariance for \uaps{}.

    \section{Quantifying Reliance on Non-Robust Features}
    \label{sec:signal}
    So far, we have shown that \uaps{} possess desirable properties that standard adversarial perturbations lack,
that make them more closely aligned with human priors.
As human priors are believed to be important for generalization, we might expect that \uaps{} contain stronger signal for generalization than standard adversarial perturbations.
Here, we find that despite their human-aligned properties, \uaps{} actually contain much less signal than standard adversarial perturbations.

We formalize this using the \emph{non-robust features} model of \citet{ilyas2019adversarial}.
We find that \uaps{} do indeed use non-robust features (that is, they modify useful correlations that models can rely on to generalize),
but the non-robust features used by \uaps{} have significantly weaker signal than the non-robust features in standard adversarial perturbations.

\subsection{The non-robust features framework}
We first review the model proposed in~\citet{ilyas2019adversarial}:
\begin{itemize}
\item A \emph{useful} feature for classification
is a function that is (positively) correlated with the correct label in expectation.
Intuitively, a feature can be thought of as computing some property of the input,
such as its color.
\item A feature is \emph{robustly useful} if, even under adversarial
perturbations (within a specified set of valid perturbations $\Delta$),
the feature is still useful.
\item A \emph{useful, non-robust} feature
is a feature that is useful but not robustly useful.
These features are useful for classification in the standard setting, but can hurt accuracy in the adversarial setting (since their correlation with the label can be reversed).
For conciseness, throughout this paper we refer to such features simply as \emph{non-robust features}.

\end{itemize}
Finally, we refer to the non-robust features leveraged by standard adversarial perturbations and \uaps{} as general non-robust features and universal non-robust features, respectively.

\subsection{Two non-robust datasets: $\Dadv$ and $\Duniv$}
\label{sec:gen_exp}
\citet{ilyas2019adversarial} show that standard adversarial perturbations leverage non-robust features, and these features alone contain enough signal for generalization.
They demonstrate this by constructing a dataset where the only features correlated with the label are those introduced by adversarial perturbations. Specifically,
they construct a dataset $\Dadv$ (which they call $\Drand$) which is modified from the original dataset as follows:
\begin{enumerate}
\item For each input-label pair $(x,y)$, select a target class $t$ uniformly at random;
then
\item \label{item:dadv} Compute an $\ell_2$ adversarial perturbation $\delta$ on $x$ targeted
  towards class $t$, and replace the original input-label pair $(x, y)$ with
  $(x + \delta, t)$.
\end{enumerate}
By construction, $x$ and $t$ are un-correlated and the only features correlated with $t$ are those leveraged by $\delta$.
And because $\delta$ is constrained to be within an $\ell_2$ ball, it can only yield non-robust features by def  inition.
\citet{ilyas2019adversarial} then train a network on $\Dadv$, and then evaluate its accuracy on a standard test set.
The surprising finding is that---despite the fact that none of the examples in $\Dadv$ are correctly labeled from our perspective---the resulting classifier achieves non-trivial classification accuracy on the test set.
Since the only signal comes from the non-robust features leveraged by the adversarial perturbations, this allows~\citet{ilyas2019adversarial} to conclude that the adversarial perturbations must contain a non-trivial amount of signal useful for generalization.

\begin{figure}[!htbp]
  \centering
  \includegraphics[width=0.65\linewidth]{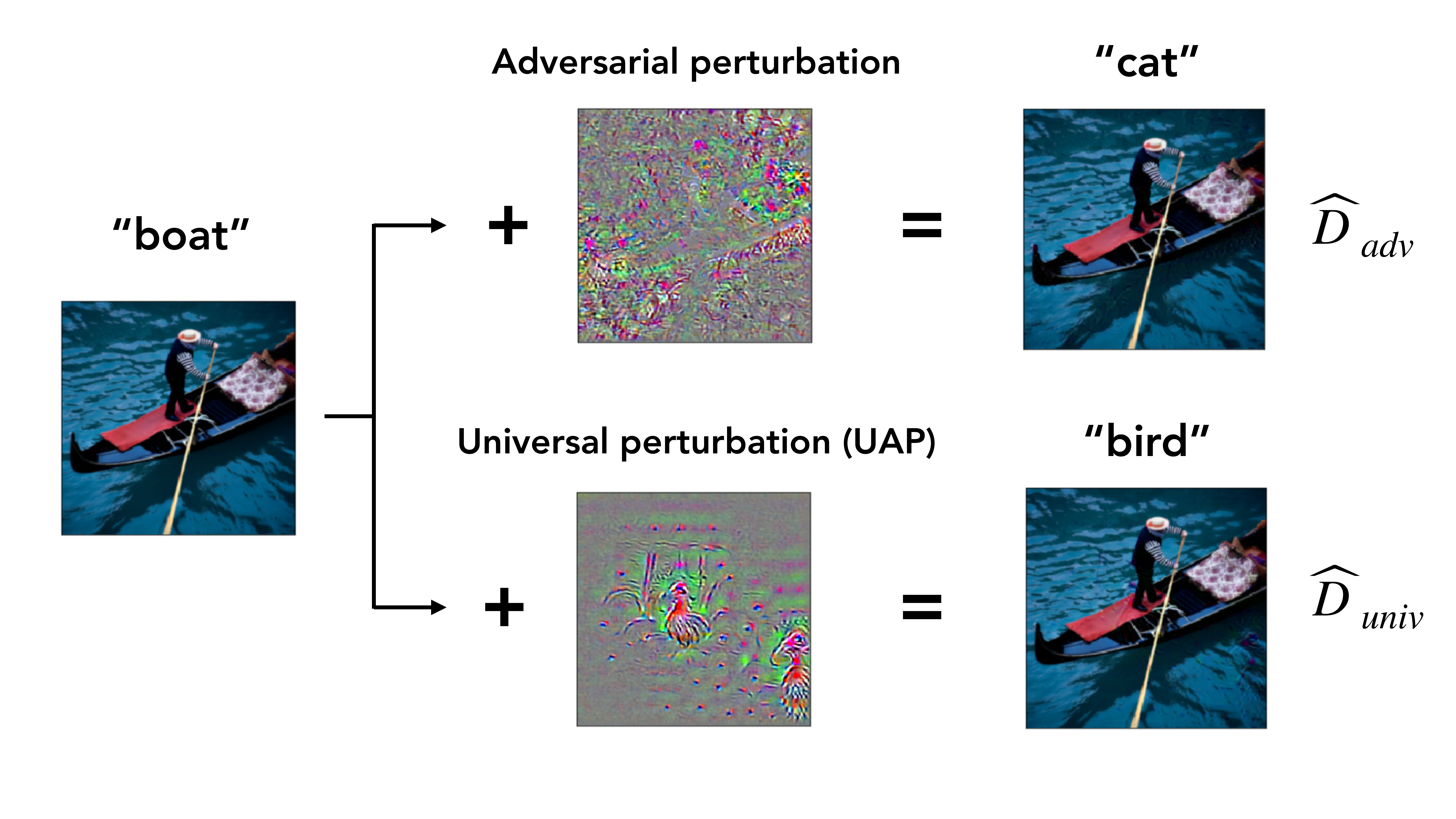}
  \caption{Construction of $\Dadv$ and $\Duniv$ datasets. The perturbed image is labeled with the target of the respective perturbation.}
  \label{fig:setup}
\end{figure}

\paragraph{Generalization from universal non-robust features.}
\label{sec:d_rand}
\label{sec:generalization}
To measure the amount of signal contained in \uaps{}, we perform the same experiment as above, except we replace the standard adversarial perturbations in Step~\ref{item:dadv} with \uaps{}.
We call the resulting dataset $\Duniv$.
We visualize this process in~\Cref{fig:setup}.
The only difference from the previous setup is that the features that remain correlated with $t$ are those leveraged by \uaps{}.
As long as the test accuracy on the (unaltered) test set of a model trained on $\Duniv$ is non-trivial, we can conclude (similarly to~\citet{ilyas2019adversarial}) that \uaps{} are leveraging non-robust features.\footnote{There is a possibility of \emph{robust feature leakage} \citep{goh2019leakage},
which is when generalization on these constructed datasets are primarily relying on robust features.
This can happen as small perturbation cannot entirely flip the correlation of a robust feature, it can still induce a small correlation by perturbing the feature. We provide analyses in \Cref{app:leakage} to bound robust feature leakage in a few ways.}
By comparing the resulting test accuracy to that achieved on $\Dadv$, we can gauge the relative strength of
generalizable signal contained in \uaps{} compared to standard perturbations.

Replicating the original setup of $\Dadv$ for $\Duniv$ requires a few additional considerations. In~\Cref{app:signal_design}, we discuss them in detail.

\paragraph{Results.} We train new ResNet-18 models on the $\Dadv$ and $\Duniv$ datasets and evaluate them on the original test set.
The best generalization accuracies from training on $\Duniv$ and $\Dadv$ were
23.2\% and 74.5\%, respectively.\footnote{We report the best results found over a grid of training hyperparameters.}
This indicates that universal non-robust features do have signal that models can use to generalize, but
universal non-robust features are harder to generalize from than general non-robust features.
Thus, there is some useful signal in universal non-robust features, but there appears to be less of it than in standard adversarial perturbations.

\subsection{Transferability of \uaps{}}
Another way to quantify the extent to which \uaps{} leverage non-robust features is by studying their transferability.
\citet{ilyas2019adversarial} suggest that the pervasive transferability of adversarial perturbations \citep{papernot2016transferability, moosavi2017universal}
can be attributed to different models relying on common non-robust features.
Consequently, perturbations that leverage more shared non-robust features
should transfer better across models. Here, we are primarily interested in
transferability as a proxy for detecting shared non-robust features, and we
compare the transferability of standard and \uaps{}.

 \begin{figure}[!hbtp]
  \centering
  \includegraphics[width=\linewidth]{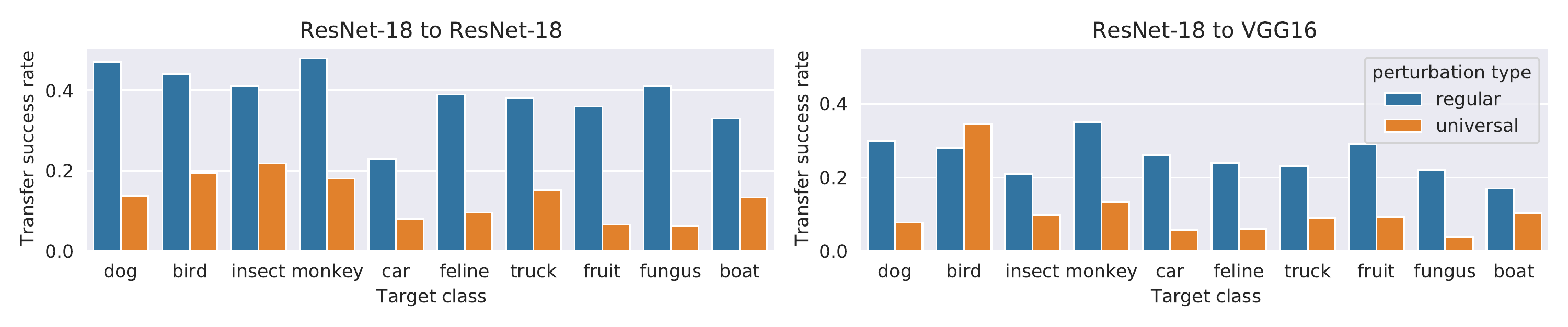}
  \caption{Comparing transferability of standard adversarial and universal adversarial
  $\ltwo$ perturbations ($\epsilon=6.0$): perturbations are generated on the source
model (ResNet-18), then transferred to a different target model: \textbf{(left)} another ResNet-18
model trained from a different random initialization, and \textbf{(right)} a VGG16 model.
Standard adversarial perturbations have a higher transfer success rate across all classes.
}
\label{fig:transfer}
\end{figure}

To measure transferability, we (1) perturb examples using either a standard adversarial
perturbation or a \uap{} on the \emph{source} model, and (2)
measure the probability that the perturbed input is classified as the target
class on a new \emph{target} model (an independently trained ResNet-18 and VGG16 model).
For normalization, to make standard and \uaps{} comparable,
we only consider perturbed images that are misclassified by the source model, and also evaluate transfer on images that have
labels different from the target label.
Figure~\ref{fig:transfer} shows that
\uaps{} transfer much less effectively to both target models, suggesting that universal non-robust features are
shared across models to a smaller extent than general non-robust features.
In~\Cref{app:signal_details}, we provide additional analyses for results in this section.

    \subsection{Interpolating Universality}
    \label{sec:interp}
    The above experiments demonstrate that while \uaps{}
are more human-aligned, they leverage only a small fraction of the statistical signal in general non-robust features.
To explore the underlying trade-off, we investigate to what extent one can
interpolate between the properties of universal and standard non-robust
features. We explore two different methods of ``interpolating''
universality: varying the number of images (\Cref{sec:interp_size}) and
varying the diversity of images (\Cref{sec:interp_class}).

\paragraph{\uaps{} over smaller base sets.}
\label{sec:interp_size}
One natural way to interpolate universality is to vary $K$, the size of the
base set (cf. \Cref{sec:prelim_computing}).  $K=1$ corresponds to standard adversarial
perturbations, and large enough $K$ corresponds to fully universal
perturbations.
We generate \uaps{} over different choices of $K$, and repeat the experiment from \Cref{sec:gen_exp}.
Table~\ref{tab:d_rand_interp} shows that signal for generalization generally
decreases with the size of the base set.
Notably, generalization begins to
suffer even for relatively small values of $K$. For example, the generalization
accuracy falls from $74\%$ at $K=1$ to just $34\%$ at $K=16$.

\begin{table}
    \parbox{.45\linewidth}{
    \centering
    \caption{Interpolating universality: generalization from $\Duniv$ constructed using \uaps{} generated with smaller base sets.}
    \label{tab:d_rand_interp}
    \begin{tabular}{rr}
    \toprule
     Size of base set ($K$) &  Test Accuracy (\%) \\
    \midrule
     1 &  74.5 \\
     2 &  57.1 \\
     4 &  61.3 \\
     8 &  57.4 \\
     16 &  34.3 \\
     32 &  21.8 \\
     256 &  19.1 \\
    \bottomrule
    \end{tabular}
    }
    \hfill
    \parbox{.45\linewidth}{
    \caption{Generalization from class-universal and sub-class universal non-robust features.
    }
    \label{tab:d_rand_class}
    \centering
    \begin{tabular}{lr}
    \toprule
     Source Class &  Test Accuracy (\%) \\
    \midrule
     Random & 23.2 \\
     Single Class &  23.9 \\
     Single Sub-class &  27.1 \\
    \bottomrule
    \end{tabular}
    }
\end{table}

\begin{figure}[!htbp]
  \centering
  \includegraphics[width=0.9\linewidth]{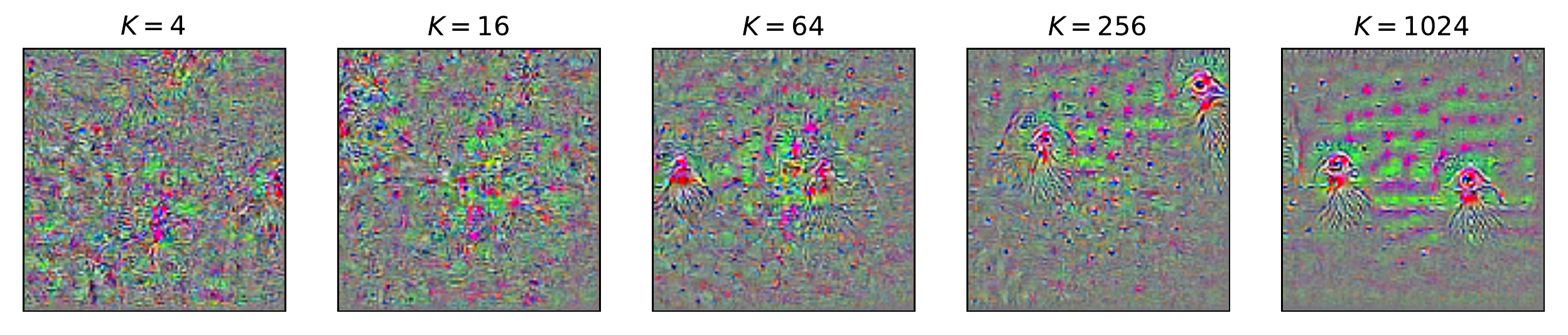}
  \caption{Visualization of \uaps{} for target \emph{bird},
  computed over base sets of different size $K$. We observe that the visual quality
  improves with the size of the base set.}
  \label{fig:vis_interp}
\end{figure}

In contrast, \Cref{fig:vis_interp} shows that the semantic quality of
\uaps{} improves as the base set becomes larger.
This suggests a possible trade-off between how human-aligned perturbations are and the amount of statistical signal, at least when interpolating along this particular axis.
However, the exact nature of this trade-off is not obvious: better semantics only become obvious at higher values of $K$ ($\gtrsim 64$), whereas generalization suffers even at relatively small values of $K$ ($\sim 16$).
Additional analysis in~\Cref{app:budget_hyp} suggests that optimizing over multiple images has a non-linear effect,
even at small values such as $K=2$.

\paragraph{\uaps{} over images of restricted diversity.}
\label{sec:interp_class}
We next look at the influence of a different factor in our observed phenomena: the semantic diversity of images in the base set.
In particular, consider \emph{class universal} and \emph{subclass universal}
perturbations, where we restrict the base set to examples from a single class
(e.g., dog) or subclass (one original ImageNet class, e.g., Terrier).\footnote{Prior works \citep{gupta2019method, benz2020double, zhang2020cd} also explore restricting the source or target classes.} %
We repeat the experiments from Section~\ref{sec:d_rand} using
these perturbations (cf. Table~\ref{tab:d_rand_class}); the
generalization on the original test set improves only modestly.

The results of these interpolation experiments show that the large gap in signal between \uaps{} and standard perturbations persists even when the level of ``universality'' is relaxed.

    \section{Related Work}
    In this section, we discuss connections to prior related work.\footnote{For references on adversarial examples more broadly, see the references in~\Cref{sec:intro}.}
\paragraph{Universal adversarial perturbations.}
\citet{moosavi2017universal} first introduce \uaps{} for classification models. Many follow-up works explore different methods to generate them, using
data-independent algorithms \citep{mopuri2017fast, liu2019universal},
singular vectors of model Jacobians \citep{khrulkov2018art},
generative models \citep{poursaeed2018generative, hayes2019learning},
and other techniques \citep{wu2019universal}.
A priori, it is not obvious why \uaps{} are so prevalent; \citet{moosavi2017analysis} shows via theory and experiments that the prevalence of \uaps{} is related to the curvature of the decision boundary.
Beyond image classification, other works \citep{hendrik2017universal, mummadi2019defending} study \uaps{} for semantic image segmentation. \citet{hendrik2017universal} find that \uaps{} generated for a fixed target segmentation exhibit local structures that resemble the target scene, analagous to the observations made here.

\paragraph{Robustness and \uaps{}.}
While the majority of research on adversarial robustness focuses on standard perturbations, prior works also study the robustness of models to \uaps{}.
Models trained via standard adversarial training \citep{madry2017towards}
are known to be more robust to \uaps{} than standard models \citep{mummadi2019defending}. \citet{mummadi2019defending} explore this further and uses \emph{shared adversarial training}, which computes \uaps{} on the mini-batch and uses them for adversarial training. Further, they study \uaps{} computed on these robustly trained models, and find that they are much more perceptible than \uaps{} computed on standard models.
\citet{shafahi2020universal} take a different approach to training, maintaining a global UAP and optimizing it in parallel with model parameters via alternating SGD.

\paragraph{\uaps{} and features.}
A few prior works also study \uaps{} from the perspective of understanding their features.
Using ideas from analysis of universal perturbations, \citet{jetley2018friends} find that class-specific patterns unintelligible to humans can induce misclassification while also being essential for classification.
\citet{zhang2020understanding} also study the \uaps{} and their features by analyzing correlations in the representation space. Their analysis contrasting \uaps{} and standard adversarial perturbations is complementary to ours; for instance, our results on spatial invariance is consistent with their finding that \uaps{} behave as the ``dominant'' feature when added to images.

\paragraph{Non-robust features.}
\citet{tsipras2019robustness} first formalize non-robust features via a simple theoretical model and study the tradeoffs between robustness and accuracy.
\citet{ilyas2019adversarial} develop these ideas further and demonstrate that non-robust features alone are sufficient for generalization on standard image classification datasets. In particular, their analysis shows that adversarial examples are a consequence of misalignment between non-robust features and human priors.
Conversely, several works discuss the usefulness and
perceptual alignment of robust features \citep{tsipras2019robustness, santurkar2019image},
which can be extracted from adversarially robust models.

\paragraph{Feature visualizations.}
Many works propose saliency map methods to construct visualizations of
features learned by networks \citep{simonyan2013deep, ribeiro2016why, sundararajan2017axiomatic, carter2019activation}.
While many of these maps are visually appealing, they
often rely on post-processing, and their visual appeal can be misleading \citep{adebayo2018sanity}.
In contrast, our visualizations of \uaps{} involve no post-processing,
yet show a striking difference between universal and standard adversarial perturbations.

    \section{Conclusion}
    In this work, we revisit (targeted) universal perturbations and show that
they exhibit human-aligned properties that distinguish them from standard adversarial perturbations.
In particular, we precisely characterize the degree to which they are human-aligned in terms of two properties:
semantic locality and spatial invariance.
We further quantify the degree to which UAPs leverage non-robust features through experiments on their generalizability and transferability,
and find that \uaps{} contain much weaker signal than standard perturbations.
Our study demonstrates that examining UAPs may be a fruitful direction for understanding
more fine-grained properties of adversarial perturbations, and associated phenomena such as the prevalence and the nature of non-robust features.

    \section*{Acknowledgements}
    Work supported in part by the NSF grants CCF-1553428 and CNS-1815221. This material is based upon work supported by the Defense Advanced Research Projects Agency (DARPA) under Contract No. HR001120C0015.

    \clearpage
    \bibliographystyle{plainnat}
    \bibliography{bib}

\clearpage
\appendix

\section*{Appendices}
\Cref{app:exp_setup} describes the experimental setup and selection of all hyperparameters.
\Cref{app:symmetry,app:signal_details,app:scaling,app:vis,app:locality,app:spatial,app:diversity} give more detailed analyses and additional results:
\Cref{app:symmetry} discusses intuition for the translational invariance observed in \Cref{sec:spatial}.
\Cref{app:signal_details} provides more results and analyses for generalization experiments in \Cref{sec:signal}.
\Cref{app:scaling} provides an alternate scaling analysis to differentiate robust and non-robust signals.
\Cref{app:vis,app:locality,app:spatial} gives a selection of additional results for experiments for~\Cref{sec:properties} across different settings.
\Cref{app:diversity} discusses the diversity of universal perturbations.

\section{Experimental Setup}
\label{app:exp_setup}
\subsection{Datasets}
\label{app:dataset}
Our experiments use CIFAR-10 and ImageNet-M10, a subset of ImageNet ILSVRC2012 \citep{russakovsky2015imagenet}; we focus primarily on ImageNet-M10.

\begin{figure*}[!htbp]
  \centering
  \includegraphics[width=\linewidth]{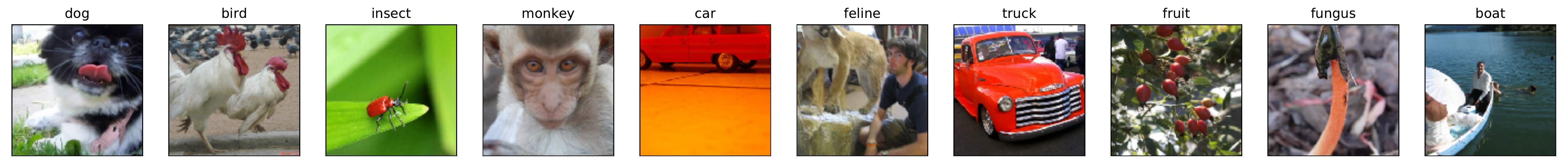}
  \caption{Sample images from each (super) class of ImageNet-M10.}
  \label{fig:m10_samples}
\end{figure*}

ImageNet-M10 consists of ten super-classes,
each corresponding a WordNet \citep{miller1995wordnet} ID in the hierarchy.
Different super-classes contain varying number of ImageNet classes, so
the dataset is balanced by choosing six classes within each super-class.
The class numbers of the corresponding ImageNet classes are shown in~\Cref{tab:m10_classes}.
The super-classes correspond to the ten labels for the classification task; for experiments using fine-grained labels,
we use the 60 ImageNet classes as labels. A sample image from each class is shown in \Cref{fig:m10_samples}.

\begin{table*}[!ht]
    \centering
    \caption{ImageNet classes used in the ImageNet-M10 dataset.}
    \label{tab:m10_classes}
    \begin{tabular}{ccc}
    \toprule
     \textbf{Class} & \textbf{WordNet ID} &\textbf{Corresponding ImageNet Classes} \\
    \midrule
     ``Dog'' & \texttt{n02084071} & 151 to 156 \\
     ``Bird'' & \texttt{n01503061} & 7 to 12 \\
     ``Insect'' & \texttt{n02159955} & 300 to 305 \\
     ``Monkey'' & \texttt{n02484322} & 370 to 375 \\
     ``Car'' & \texttt{n02958343} & 407, 436, 468, 511, 609, 627 \\
     ``Feline'' & \texttt{n02120997} & 286 to 291 \\
     ``Truck'' & \texttt{n04490091} & 555, 569, 675, 717, 734, 864 \\
     ``Fruit'' & \texttt{n13134947} & 948, 984, 987 to 990 \\
     ``Fungus'' & \texttt{n12992868} & 991 to 996 \\
     ``Boat'' & \texttt{n02858304} & 472, 554, 576, 625, 814, 914 \\
    \bottomrule
    \end{tabular}
\end{table*}

The ImageNet-M10 dataset can be used easily on top of a standard ImageNet directory by
loading the \texttt{mixed\_10} dataset from the \texttt{robustness} library \citep{robustness}.

ImageNet-M10 is comparable in size and complexity to restricted ImageNet, which has been used for adversarial robustness research \citep{ilyas2019adversarial, santurkar2019image}.

\subsection{Hyperparameters}
\label{app:hyper}

All experiments in the paper involve a combination of: training models,
computing standard adversarial perturbations, and computing \uaps{}.
Below, we provide details for all the hyperparameters and their selection.

\textbf{Robustness threshold}
We use the same robustness threshold throughout: $\ltwo, \epsilon=6.0$ and $\linf, \epsilon=8/255$ for ImageNet datasets, and $\ltwo, \epsilon=1.0$ for CIFAR-10.

\subsubsection{Training models}
\textbf{Training base models}
All models are trained with SGD with momentum and standard data augmentation. For corresponding robust models\footnote{These are only used for the analysis in~\Cref{app:scaling}.}, the same optimization parameters are used with projected gradient descent (PGD) \citep{madry2017towards} using 3 steps and attack step size $\frac{2}{3}\epsilon$ (for both $\ltwo$ and $\linf$).
Weight decay of $5\cdot{10}^{-4}$ was used in all cases.
The accuracies of these models are shown in \Cref{tab:models}.

\begin{table*}[!ht]
    \centering
    \caption{Standard and robust models (ResNet-18) used in our experiments.}
    \label{tab:models}
    \begin{tabular}{cccc}
    \toprule
     \textbf{Dataset} & \textbf{Model} &  \textbf{Standard Test Accuracy (\%)} &
     \textbf{Robust Test Accuracy (\%)} \\
    \midrule
     ImageNet-M10 & Standard & 95.7 & $<1.0$ \\
                  & $\ltwo$  & 86.7 & 59.8 \\
                  & $\linf$  & 87.6 & 59.2  \\
    \midrule
     CIFAR-10     & Standard & 94.8 & $<1.0$  \\
                  & $\ltwo$  & 80.0 & 50.7 \\
    \bottomrule
    \end{tabular}
\end{table*}

\textbf{Training models on constructed datasets}
The models are trained on the constructed datasets, including $\Duniv$ and $\Dadv$ (Section~\ref{sec:d_rand}), over the following grid of hyperparameters:
three learning rates (0.01, 0.05, 0.1), two batch sizes (128, 256), including/ not including a single learning rate drop by a factor of 10. All models are trained with 400 epochs, standard data augmentation,\footnote{We use standard ImageNet augmentations: random crop, horizontal flip, color jitter, and rotation. Without data augmentation, models overfit these constructed datasets.} and weight decay $5 \cdot 10^{-4}$.
We report the setting corresponding to the highest test accuracies in \Cref{tab:hparams}.

As an exception, for the $K$-interpolated datasets (Section~\ref{sec:interp}), we fixed a single set of hyperparameters to train the models.

\begin{table*}[!ht]
    \centering
    \caption{Default and best hyperparameters for training models on different datasets.}
    \label{tab:hparams}
    \begin{tabular}{llccccc}
    \toprule
    \textbf{Source Dataset} & \textbf{Constructed Dataset} &  \textbf{Epochs} & \textbf{LR} & \textbf{Batch Size} & \textbf{LR Drop} \\
    \midrule
     ImageNet-M10 & $\mathcal{D}$ (original) & 200 & 0.1 & 256 & 50, 100, 150 \\ %
                  & $\Dadv$ ($\ltwo$)  & 400 & 0.01 & 128 & 250 \\ %
                  & $\Duniv$ ($\ltwo$, random class) & 400 & 0.01 & 128 & 250 \\ %
                  & $\Duniv$ ($\ltwo$, $K$-interp)  & 400 & 0.01 & 256 & 250 \\ %
                  & $\Duniv$ ($\ltwo$, same class)  & 400 & 0.05 & 64 & None \\ %
                  & $\Duniv$ ($\ltwo$, same subclass)  & 400 & 0.05 & 128 & 250 \\ %
                  & $\Dadv$ ($\linf$)  & 400 & 0.001 & 256 & None \\ %
                  & $\Duniv$ ($\linf$) & 400 & 0.1 & 128 & None \\ %
     CIFAR-10     & $\mathcal{D}$ (original) & 150 & 0.1 & 128 & 50, 100 \\ %
                  & $\Dadv$ ($\ltwo$)  & 400 & 0.05 & 256 & 250 \\ %
                  & $\Duniv$ ($\ltwo$) & 400 & 0.05 & 256 & None \\ %
    \bottomrule
    \end{tabular}
\end{table*}

\subsubsection{Computing adversarial perturbations}

\textbf{Adversarial perturbations}
For a given $\epsilon$, PGD with an attack step size of $\frac{\epsilon}{3}$ and 10 steps is used (no random restarts.)

\begin{figure*}[!htbp]
  \centering
  \includegraphics[width=\linewidth]{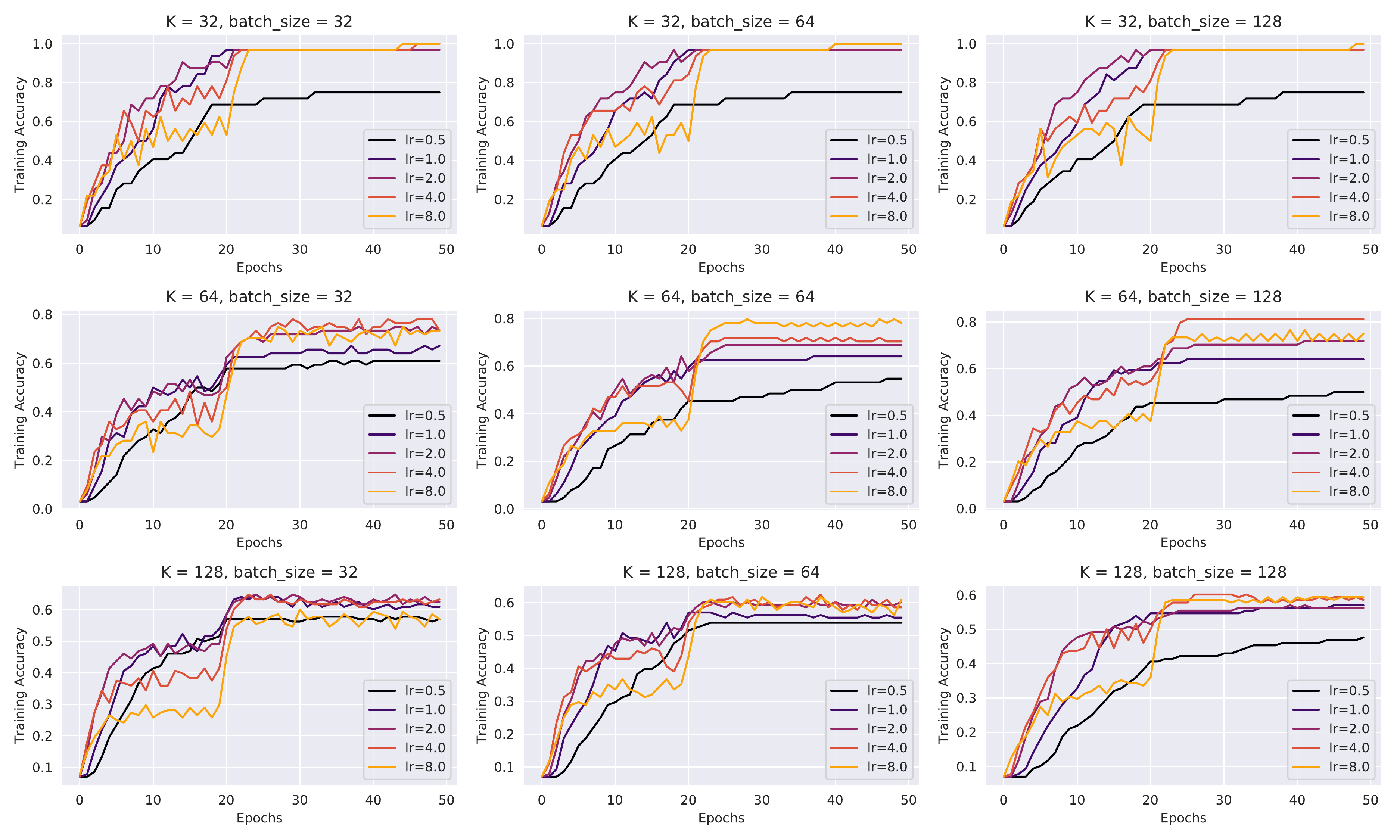}
  \caption{The learning dynamics of computing universal perturbation (epoch vs. training accuracy on the base set) are shown across a selection of learning rates, batch sizes, and three intermediate values of $K$.}
  \label{fig:learning_grid}
\end{figure*}

\textbf{Universal perturbations}
The hyperparameters of the PGD algorithm for universal perturbations (\Cref{sec:prelim_computing}) are:
learning rate schedule\footnote{Also known as \emph{step size} in the adversarial examples literature.},
number of epochs, and batch size.\footnote{Batch size only matters if it is smaller than the size of the base set, $K$.}
In order to see the impact of the choice of these parameters, we ran a grid search over the parameters; a sample of these selections is visible in Figure~\ref{fig:learning_grid}.
Overall, we observe that the exact choice of these parameters did not have a large impact on the final training accuracy (and the quality of the resulting perturbations, both in terms of visuals and generalization to other images) as long as (1) the initial learning rate is not too low or too high, and (2) a sufficient number of epochs is used.
So in all of our experiments, we fix a learning rate of 2.0 and a batch size of 128 or 256, and trained a sufficient number of epochs (max of 100).\footnote{In this case $K$ is much larger than batch size, fewer epochs suffice as the effective number of updates increases due to bounded batch size. In particular, for a fully universal base set (e.g. the entire training set), as few as five epochs seed to suffice.} The use of a learning rate decay did not seem to have much impact, so we use the same learning rate throughout.

\subsection{Computing resources}
Experiments were run primarily on 8 NVIDIA GeForce GTX 1080 Ti GPUs.

\section{Invariance of UAPs}
\label{app:symmetry}
Intuitively, it may not be surprising that UAPs are translationally invariant,
given the input-independence of UAPs.
We sketch an argument that formalizes this intuition.

First, by definition, a universal perturbation is (with high probability) effective against both image $x$,
and its translation $T(x)$, where $T$ is a translation operation from the group of translations (with wrap-around), as we expect
$T(x)$ as a sample from the same underlying distribution.\footnote{Note we compute universal perturbations over a fixed set of images, without data augmentation, so we do not directly introduce translational invariance.}
Now, consider the image $T(x) + \delta$ where $\delta$ is a universal perturbation.
If we assume that our models are reasonably translationally invariant,
it will classify $T(x) + \delta$ similarly as $T^{-1}(T(x) + \delta) = x + T(\delta)$, where $T^{-1}$ is the inverse translation.\footnote{The equality assumes associativity, and we are ignoring boundary effects (e.g. rounding) for simplicity.} Hence, $T(\delta)$ should also be an effective perturbation for $x$. Since $x$ was arbitrary, $T(\delta)$ is a universal perturbation.

More generally, we expect UAPs to be invariant to transformations under which our input distribution is also invariant to.

\section{Additional Analyses for \Cref{sec:signal}}
\label{app:signal_details}
\subsection{Design considerations}
\label{app:signal_design}
Replicating the original setup of $\Dadv$ for $\Duniv$ requires a few additional considerations,
due to the different nature of \uaps{}.

\textbf{Diversity of perturbations.}
Firstly, it is rather expensive to compute a
separate \uap{} for each input, so we compute and use the same
\uap{} per batch of inputs. One possible concern is that re-using same perturbations across the batch results in a
reduction in ``diversity'' of features considered. We ran additional experiments to control for this, but found that increasing the number of unique \uaps{} used only marginally increases the generalization, and does not impact any of the global trends. Further, increasing the visual diversity of perturbations via other methods also had limited impact on increasing generalization (see \Cref{app:diversity} for methods to increase diversity).

\textbf{UAPs fail on some inputs}.
Another complication is that for the robustness threshold $\epsilon$ considered here, universal
perturbations fool models only on a fraction $\rho$ of all training examples (while
adversarial perturbations at the same threshold succeed on all examples). Hence, we only include
examples on which the \uap{} succeeds (i.e., a pre-trained
model classifies the perturbed input as $t$). This filtering step reduces the
size of the dataset, so we decrease the number of examples in $\Dadv$
accordingly in our evaluation. We also ensure that the
distribution of new and original labels is uniform and independent so that the
original label is uncorrelated with the new label.

\subsection{All results on generalization on non-robust datasets}
\label{app:generalization}

Here we collect additional results for the generalization experiment across various settings:
$\ltwo, \linf$ perturbations in ImageNet-M10 and $\ltwo$ perturbations on
on CIFAR-10.
We did not conduct experiments on ImageNet,
as the constructions of the datasets is computationally too intensive.

\Cref{tab:full_constructed} shows that there is consistently much lower generalization from $\Duniv$
than from $\Dadv$.
\begin{table*}[!ht]
    \centering
    \caption{Generalization from constructed datasets.}
    \label{tab:full_constructed}
    \begin{tabular}{cccc}
    \toprule
     \textbf{Source Dataset} & \textbf{Perturbation Set} & \textbf{Constructed Dataset}
      & \textbf{Test Accuracy} (\%) \\
    \midrule
     ImageNet-M10 & $\ltwo, \varepsilon=6.0$ & $\Dadv$ & 74.5 \\
     ImageNet-M10 & $\ltwo, \varepsilon=3.0$ & $\Dadv$ & 76.6 \\
     ImageNet-M10 & $\ltwo, \varepsilon=6.0$ & $\Duniv$ & 23.2 \\
     ImageNet-M10 & $\linf, \varepsilon=8/255$ & $\Dadv$ & 78.7 \\
     ImageNet-M10 & $\linf, \varepsilon=8/255$ & $\Duniv$ & 26.5 \\
     CIFAR-10 & $\ltwo, \varepsilon=1.0$ & $\Dadv$ & 64.3 \\
     CIFAR-10 & $\ltwo, \varepsilon=1.0$ & $\Duniv$ & 23.3 \\
    \bottomrule
    \end{tabular}
\end{table*}

\subsection{Bounding robust feature leakage}
\label{app:leakage}

The accuracies on the $\Duniv$ datasets are very low (for comparison, they are even lower
compared to a ResNet-18 model trained on (fixed) randomly initialized features, which achieve an accuracy greater than 30\%).
Given this, one concern is that the small signal we observe can be entirely accounted for by \emph{leakage of robust features} into these datasets \citep{ilyas2019adversarial, goh2019leakage}:
while a small perturbation cannot entirely flip the correlation of a robust feature, it can still induce a small correlation on average on the $\Duniv$ dataset.
Even if there are no non-robust features that are perturbed, the small ``leaked'' correlations from the robust features
alone could allow for generalization from $\Duniv$ to the original test set.

We run several additional experiments to bound the amount of leakage
(we focus on $\linf$ perturbations on ImageNet-M10):
\begin{itemize}
\item We construct a dataset similar to $\Ddet$ \citep{ilyas2019adversarial} but instead using universal perturbations,
where we choose new corrupted labels by cyclically shifting original labels, rather than choosing them randomly; the robust features now point \emph{away} from the label.
Models trained on this new dataset achieve an accuracy up to 19.1\% on the original test set, which is less than 26.5\% from $\Duniv$, but still well above chance-level of 10\%.
This demonstrates that there is residual signal in universal perturbations that cannot be entirely accounted for by robust leakage.

\item We can also probe the original dataset by seeing how well different features\footnote{As usual, we consider the penultimate layer representations of the network before the final linear classifier.} can fit the dataset.
Adapting the procedure in \citet{goh2019leakage}, we take pre-trained features
from different natural and $\linf$ robust models on ImageNet-M10, and train a linear classifier over those features on $\Duniv$. The results in \Cref{tab:leakage} show that features from natural models,
even sourced from a different model than the one used to generate $\Duniv$ or from a different architecture,
capture more of the signal in $\Duniv$ than robust features.
For comparison, the model trained on $\Duniv$ from scratch, as we saw earlier, achieves an accuracy of 26.5\%,
which is higher than 22.3\%.
This demonstrates again that while some leakage is happening, it cannot account for all of the signal
in universal perturbations.

\begin{table*}[!ht]
    \centering
    \caption{Training on $\Duniv$ using fixed features to check for leakage. The gap between the last row (robust) and others suggest that there are indeed non-robust features being utilized beyond (leaked) robust features.}
    \label{tab:leakage}
    \begin{tabular}{cc}
    \toprule
     Feature Source & (Fine-tuned) Test Accuracy (\%) \\
    \midrule
     natural &  36.9 \\
     natural, diff init & 35.0 \\
     natural, diff arch (VGG16) & 26.4 \\
     robust ($\linf$) & 22.3 \\
    \bottomrule
    \end{tabular}
\end{table*}

\end{itemize}

These results consistently show that while some robust leakage occurs, it cannot account for all of the signal in universal perturbations. A small but non-trivial amount of signal comes from the universal non-robust features.

\subsection{Evidence of interaction between images in the base set}
\label{app:budget_hyp}
The degradation in the non-robust signal at even small values of $K$ is striking.
One plausible hypothesis for the degradation of signal quality between $K=1$ and $K > 1$ is
that optimizing the perturbation over more images splits the effective perturbation norm budget,
and is akin to optimizing over each image separately with a reduced budget, say $\epsilon / K$.
To test this hypothesis, we constructed standard adversarial perturbations with a reduced budget of $\epsilon=3.0$.
This corresponds to the setting $K=1, \epsilon=3.0$.
We compare the generalization from this new dataset to that from the $K=2, \epsilon=6.0$ case in \Cref{tab:budget_hypothesis}
(latter is the same value as in \Cref{tab:d_rand_interp}), which should be similarly effective if the
norm budget hypothesis is correct.
We instead find that the standard adversarial perturbation at a smaller norm of $\epsilon=3.0$ results in much hire generalization accuracy.
Given the large gap, it seems unlikely that this simple norm budget sharing can explain the above interpolation phenomena.
This suggests that when universal perturbations are computed jointly over multiple images,
the perturbations are of a fundamentally different nature.
This may be due to the perturbation having to interact with different non-robust features present in different images.

\begin{table}[!ht]
    \centering
    \caption{Comparing generalization from two different constructed datasets for ImageNet-M10 using $\ltwo$ perturbations.
    $K$ is the base set size; $K=1$ corresponds to using standard adversarial perturbations, and $K=2$ to computing perturbations over random pairs of images.}
    \label{tab:budget_hypothesis}
    \begin{tabular}{cc}
    \toprule
     Perturbation Type & Test Accuracy on Original (\%) \\
    \midrule
     $\epsilon=6.0$, $K=2$ & 57.1 \\
     $\epsilon=3.0$, $K=1$ & 76.6 \\
    \bottomrule
    \end{tabular}
\end{table}

\subsection{Alternate base sets}
In this paper, we focused on computing universal perturbations over base sets (cf. \Cref{sec:prelim_computing})
consisting of different images (from possibly a restricted set of classes).
To study the importance of sample diversity in the base set, we computed universal perturbations
over the base set consisting of $K$ different augmentations of a \emph{single} image.
The resulting universal perturbations still have similar (but less salient) visual characteristics, and are transferable but to a lesser degree than perturbations computed over base set of different images.

\section{Scaling Analysis}
\label{app:scaling}
Results of~\Cref{sec:signal} and ~\Cref{app:leakage} show that
universal perturbations are leveraging non-robust features.
However, they do not entirely rule out contributions from robust features,
and suggest some robust feature leakage may be occurring.
If universal perturbations are also relying on robust features, we cannot necessarily attribute the semantic aspects of the perturbation that we observed earlier (and which we showed to be responsible for most the ASR) to non-robust features.
Nonetheless, here we provide evidence that robust leakage is unlikely to be a primary contribution to the signal.

\textbf{Scaling analysis on natural vs. robust model} \\
Our analysis based on the following premise: if the universal perturbations are primarily relying on small perturbations of robust features,
then a robust model should eventually react when we amplify the signal in the universal perturbations.
We currently lack tools to selectively control robust and non-robust contents of an image within the $\ell_p$ ball, so we settle for a simpler proxy.

\begin{figure*}[!htbp]
    \centering
    \includegraphics[width=0.48\linewidth]{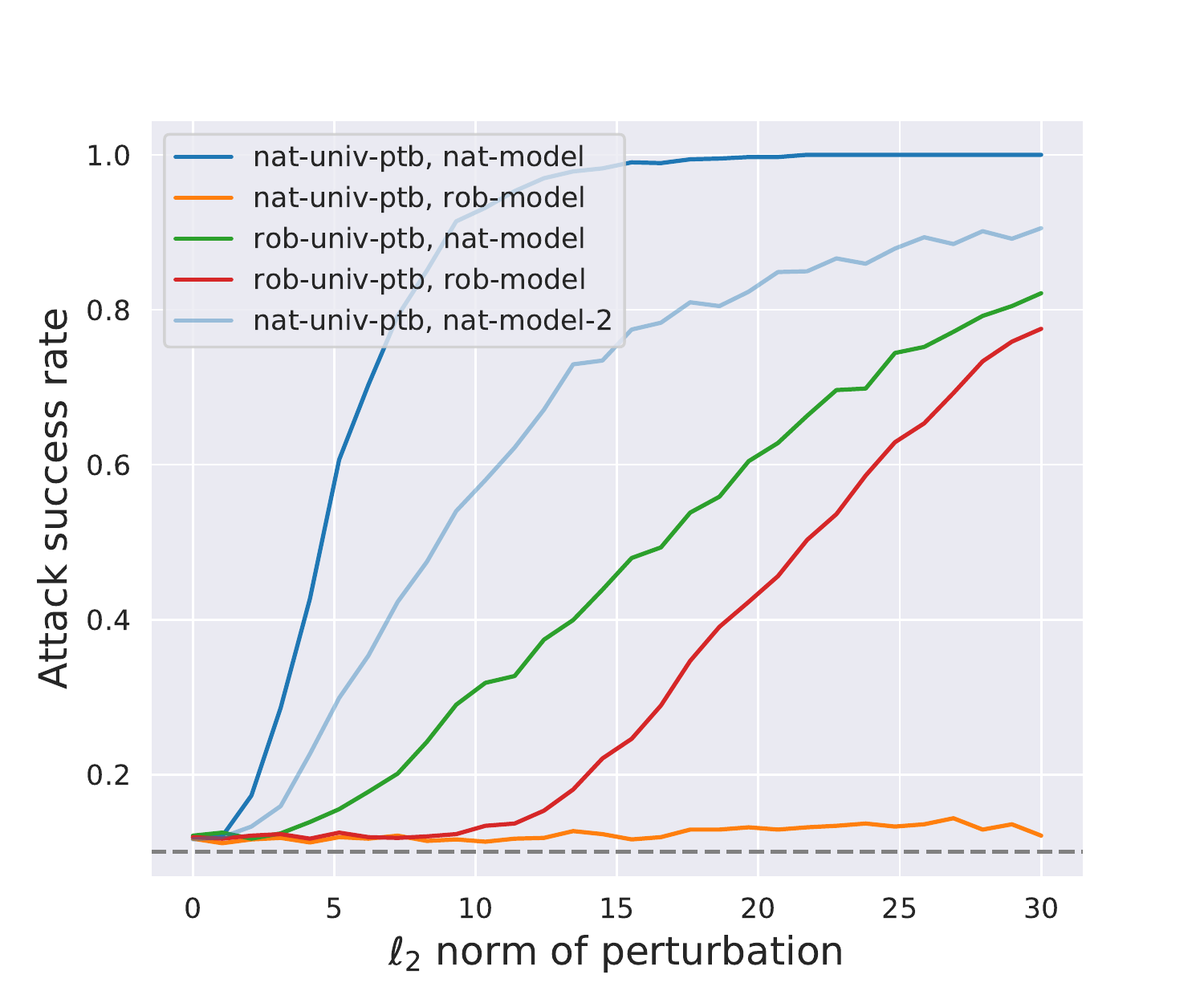}
    \includegraphics[width=0.48\linewidth]{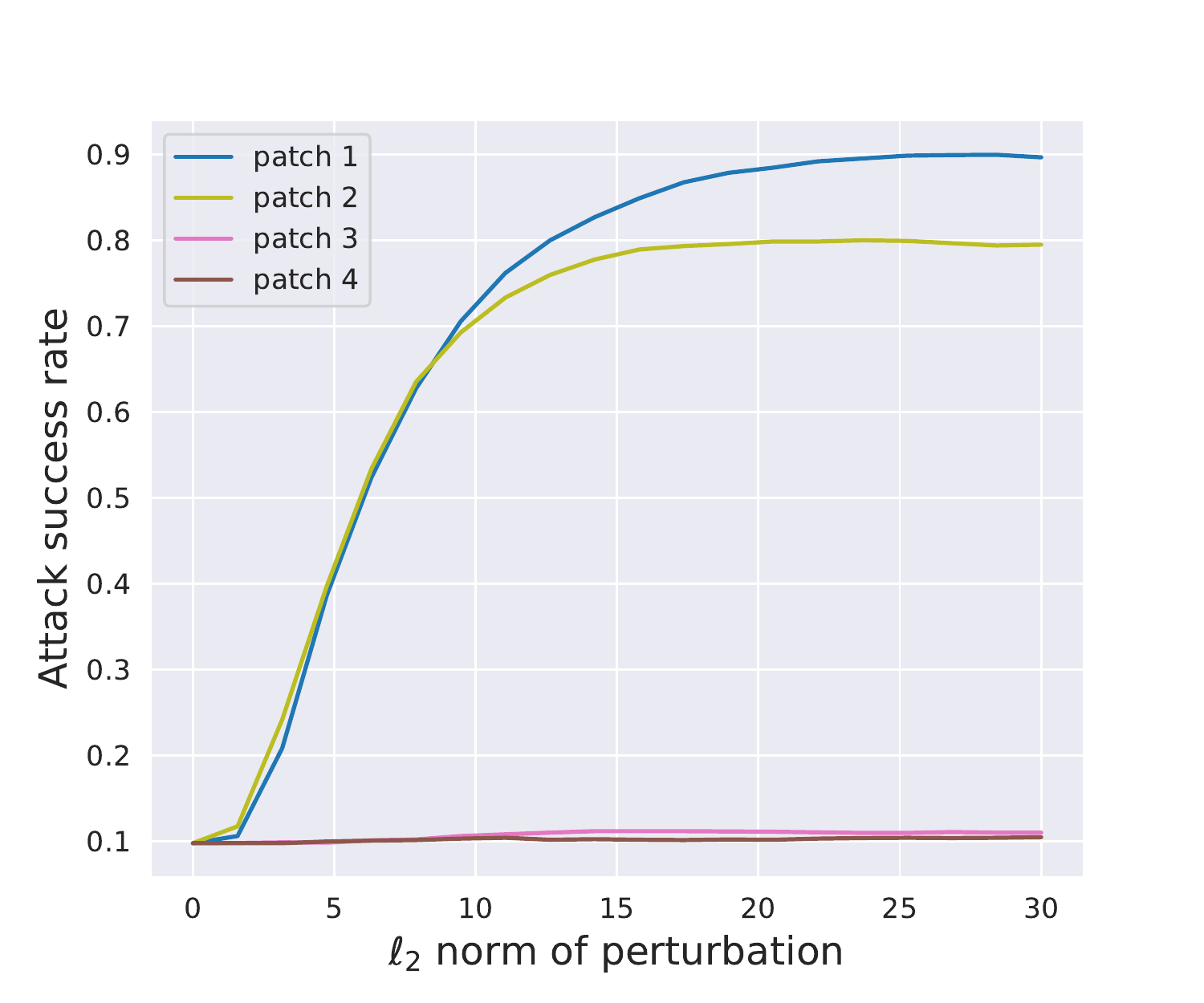}
    \caption{Scaling analysis of universal perturbations on ImageNet-M10:
    (Left) we generate a universal perturbation (\texttt{nat-univ-ptb}) of $\ltwo$ norm
    $\epsilon=6.0$ on a natural model (\texttt{nat-model}) for the target \emph{bird}, and measure the sensitivity of the natural and a robust model (\texttt{rob-model}) at various rescalings of the perturbation. We also generate a universal perturbation (\texttt{rob-univ-ptb}) of norm $\epsilon=30.0$ for the robust
    model and use it to show that the robust model \emph{does} react to scalings of
    some perturbation.  The same \texttt{nat-univ-ptb}
    is also evaluated on an independent natural model (\texttt{nat-model-2})
    for control; (Right) different cropped local patches of the same
        \texttt{nat-univ-ptb} are also evaluated on the natural model}
    \label{fig:scaling}
\end{figure*}

Below, we first show that simply linearly scaling the perturbation can effectively tune the strength of the signal.
More precisely, given a universal perturbation $\delta$, we evaluate the ASR of $t\cdot\delta$, where $t$ is a scaling parameter.\footnote{\citet{jetley2018friends} use a similar analysis to study subspaces that are essential for classification and also susceptible to adversarial perturbations.}
When we perform this scaling on a natural (non-robust) model, the ASR increases as we increase $t$ (\Cref{fig:scaling}).
(This holds even if networks used to train the universal perturbation and the one used to evaluate its ASR differ.)
If there was significant robust leakage, we would expect a similar qualitative behavior for the robust model.
Instead, we find that the robust model does \emph{not} react to any scaling of the perturbation.
On the other hand, a universal perturbation of norm $\epsilon = 30.0$\footnote{Universal perturbations on a robust model require a larger norm as the model is robust near the $\epsilon$ it is trained at.} is computed on an adversarially-trained robust model ($\epsilon = 6.0$) still fools the natural model, even when scaled down.
Together, these findings suggest that universal perturbations computed on the natural model are unlikely to leverage robust features used by the robust model.

\begin{figure*}[!htbp]
    \centering
    \includegraphics[width=0.4\linewidth]{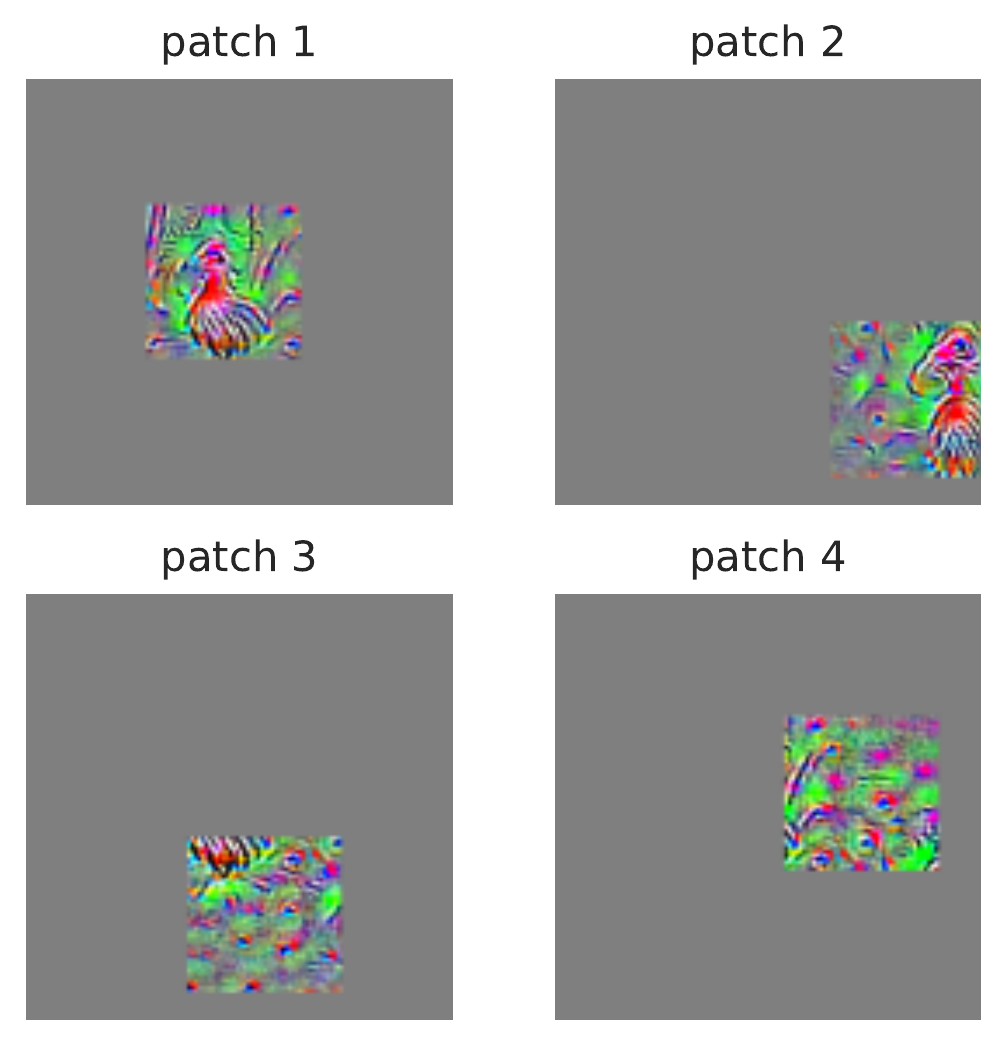}
    \caption{Visualizations of the cropped patches used in~\Cref{fig:scaling} (Right)}
    \label{fig:patches}
\end{figure*}

In fact, a more fine-grained scaling analysis shows that the signal is indeed primarily amplified in the semantic parts of the image.
In Section~\ref{sec:properties} we already observed that models are most sensitive to the most semantic parts of universal perturbations; here we check that the order remains consistent across various scales.
We repeat the above scaling evaluation on natural vs. robust models, except we scale only patches of the perturbation, of varying levels of semantic content (Figure~\ref{fig:scaling}).
As before, we find that none of the patches, even after scaling, trigger the robust network.
On the non-robust models, the relative effectiveness of local patches is consistent with their semantic content: patches that are more semantically identifiable have more signal at all scales.

The above analysis shows that universal perturbations are likely primarily targeting non-robust features. Combined with our findings earlier in \Cref{sec:properties}, this shows that non-robust features may not be entirely unintelligible to humans.
Our visualizations (\Cref{fig:patches}) suggests that
universal non-robust features may be capturing intuitive properties, such as the presence of a faint bird head.
Thus, non-robustness in networks may partially arise from
relying on cues that are small in magnitude but still human-aligned.

\section{Visualizations}
\label{app:vis}

We show visualizations of universal perturbations across different datasets (CIFAR-10, ImageNet-M, and ImageNet), architectures (VGG for Mixed10), and norm constraint ($\linf$ for Mixed10) to illustrate the generality of the phenomena observed.

\paragraph{Normalization for visualization.}
For all visualizations, perturbations were rescaled as follows (for each input channel): first truncated to
$[-3\sigma, 3\sigma]$ (where $\sigma$ is the standard deviation across all channels and pixels), then rescaled and shifted to lie in $[0,1]$.

\subsection{Different datasets}
\label{app:other_data}

\begin{figure*}[!htb]
  \centering
  \includegraphics[width=\linewidth]{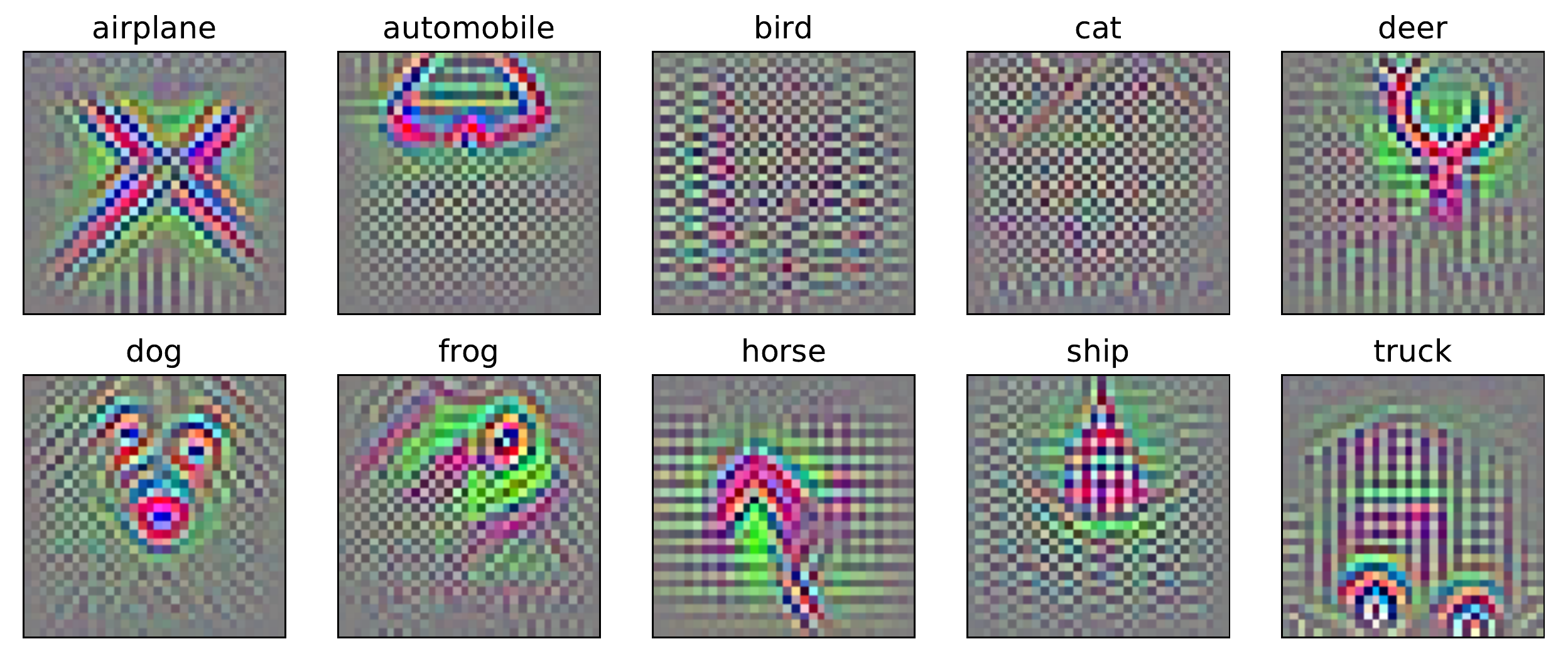}
  \caption{Visualization of $\ell_{2}$ universal perturbations at $\epsilon=1.0$ for a \textbf{CIFAR-10} model. Optimization on some classes (bird and cat) fail occasionally, and is reflected in their lack of semantic content.}
  \label{fig:cifar_univ}
\end{figure*}

\begin{figure*}[!htb]
  \centering
  \includegraphics[width=\linewidth]{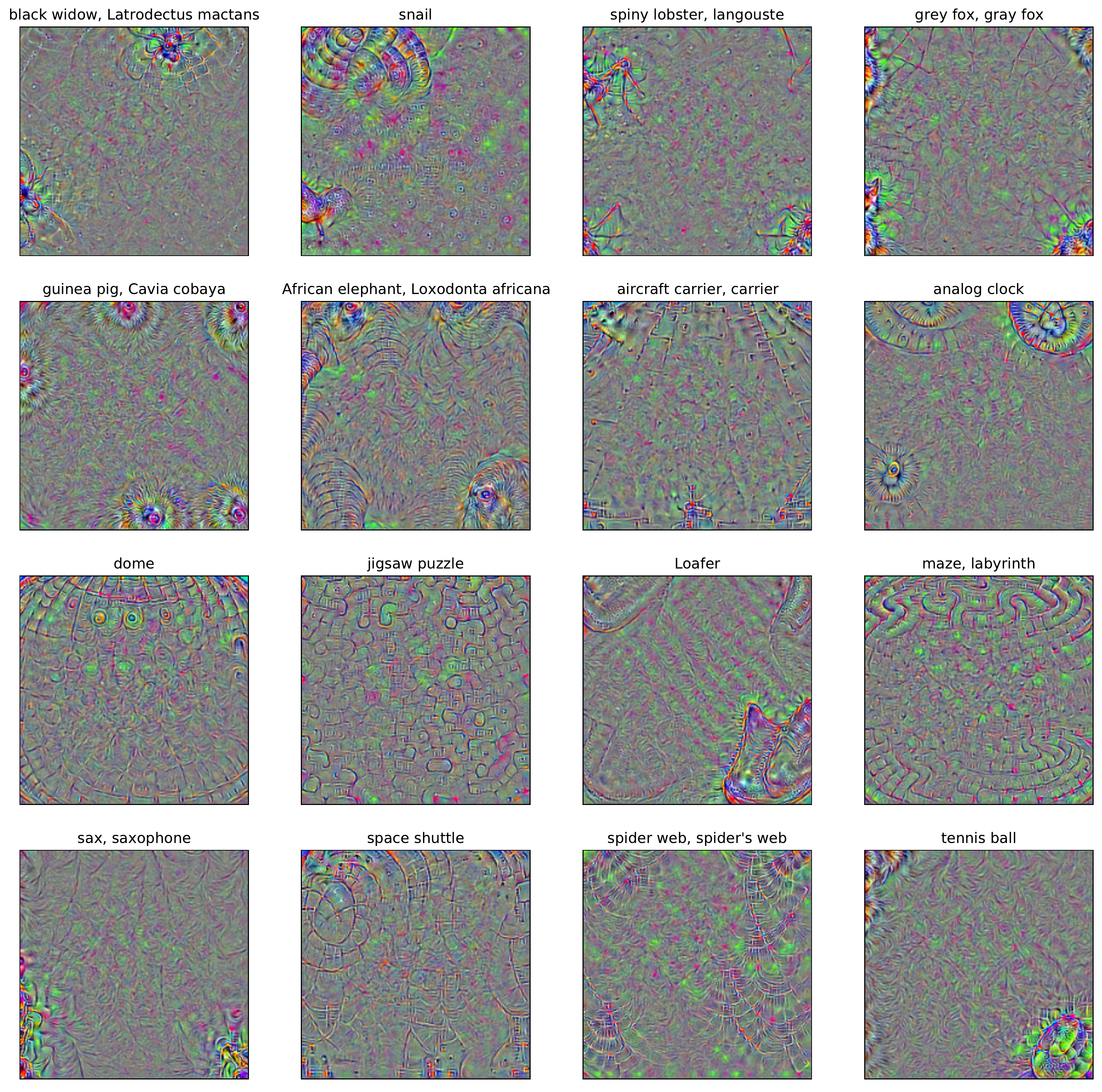}
  \caption{Visualization of $\ell_{2}$ universal perturbations at $\epsilon=6.0$ for a full \textbf{ImageNet} model. We hand-picked the classes with the most salient perturbations. Many of the perturbations possess visual characteristics identifiable with the corresponding class.}
  \label{fig:imagenet}
\end{figure*}

See~\Cref{fig:cifar_univ,fig:imagenet}.

\subsection{Other architectures}
\label{app:other_arch}
\begin{figure*}[!htb]
  \centering
  \includegraphics[width=\linewidth]{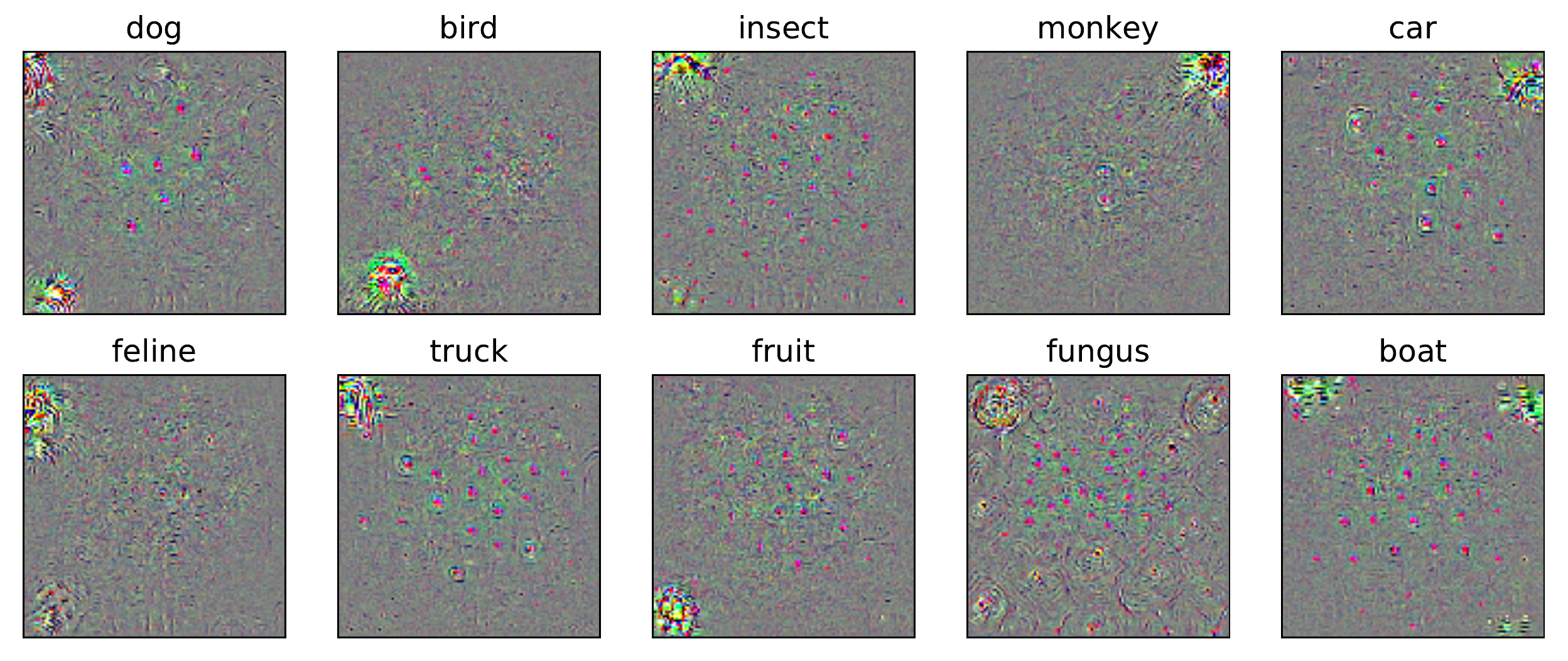}
  \caption{Visualization of $\ell_{2}$ universal ($K=512$) perturbations at $\epsilon=6.0$ for ImageNet-M10,
  generated this time on a \textbf{VGG11} network \citep{simonyan2015very}.}
  \label{fig:other_arch}
\end{figure*}

See~\Cref{fig:other_arch}.

\subsection{$\ell_{\infty}$ perturbations}
\label{app:linf}
\begin{figure*}[!htb]
  \centering
  \includegraphics[width=\linewidth]{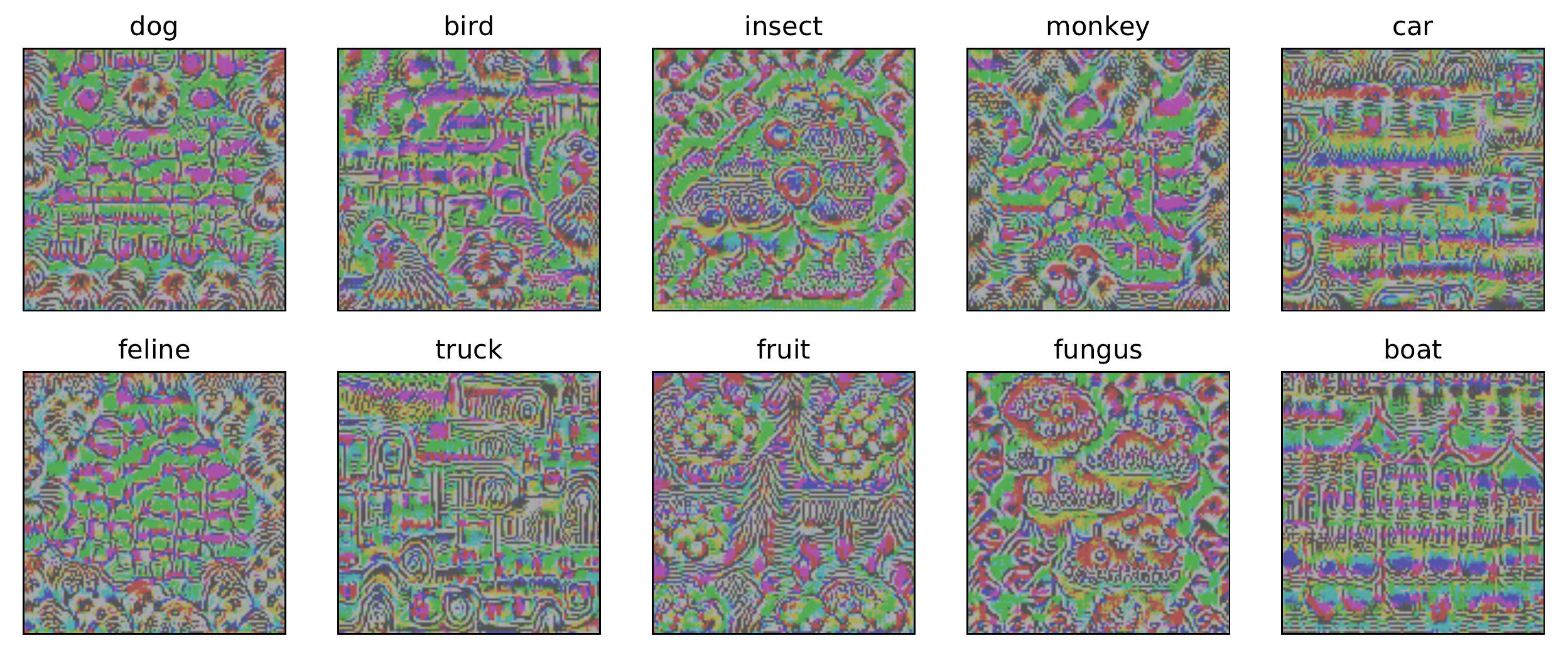}
  \caption{Visualization of $\ell_{\infty}$ universal perturbations at $\epsilon=8/255$
  for different target classes. We observe recognizable patterns for each class similar to those of $\ell_2$ universal perturbations, but the remaining regions have much higher energy and different texture.}
  \label{fig:linf_univ}
\end{figure*}

See~\Cref{fig:linf_univ}.

\section{Locality Analysis}
\label{app:locality}

\begin{figure*}[!htbp]
  \centering
  \includegraphics[width=0.8\linewidth]{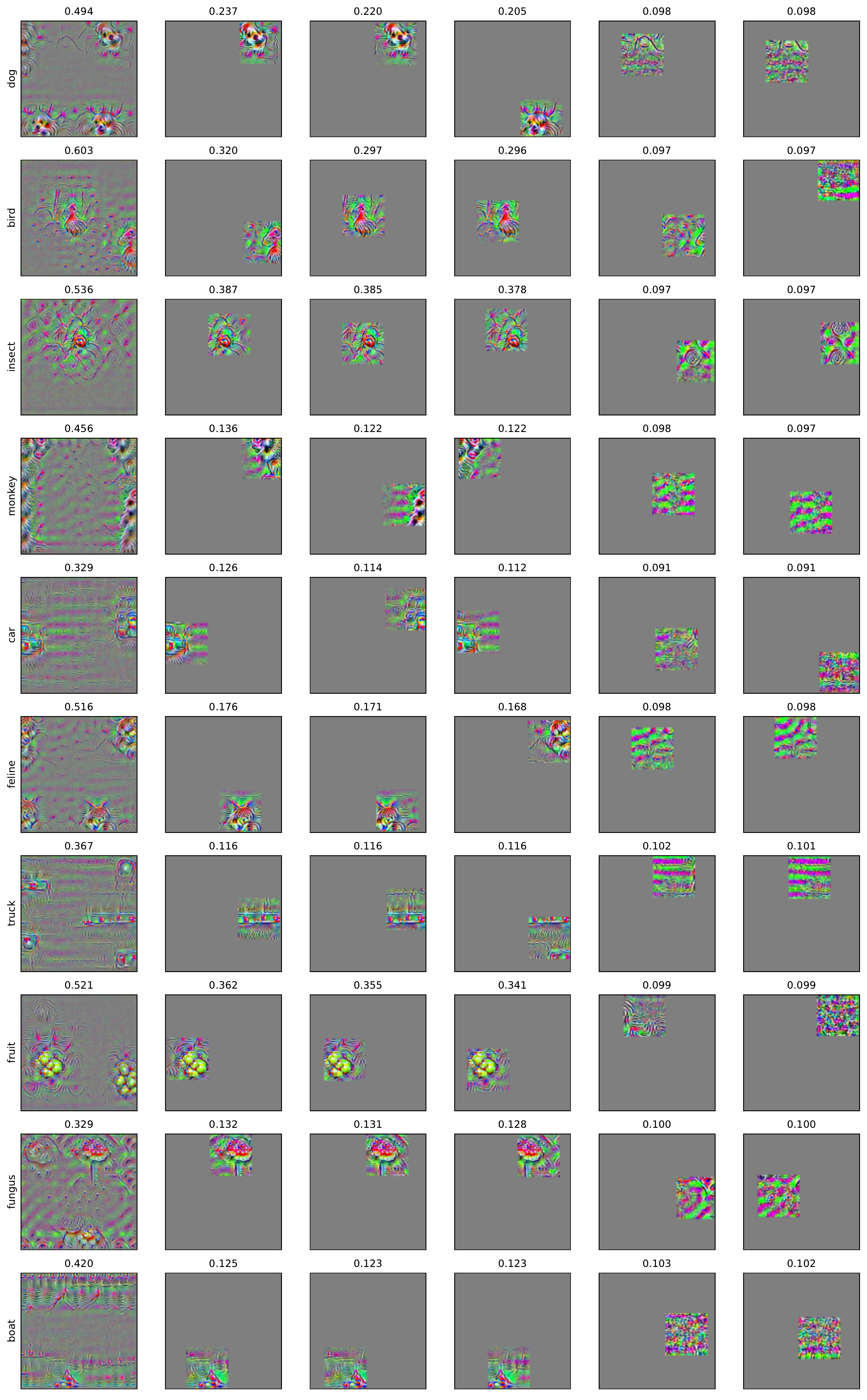}
  \caption{Locality analysis of $\ltwo$ universal perturbations on ImageNet-M10. Each row shows the perturbation for the particular class, followed by three patches with highest ASR and two patches with lowest ASR; ASR is written above each patch.}
  \label{fig:full_locality_m10_l2}
\end{figure*}

\begin{figure*}[!htbp]
  \centering
  \includegraphics[width=0.8\linewidth]{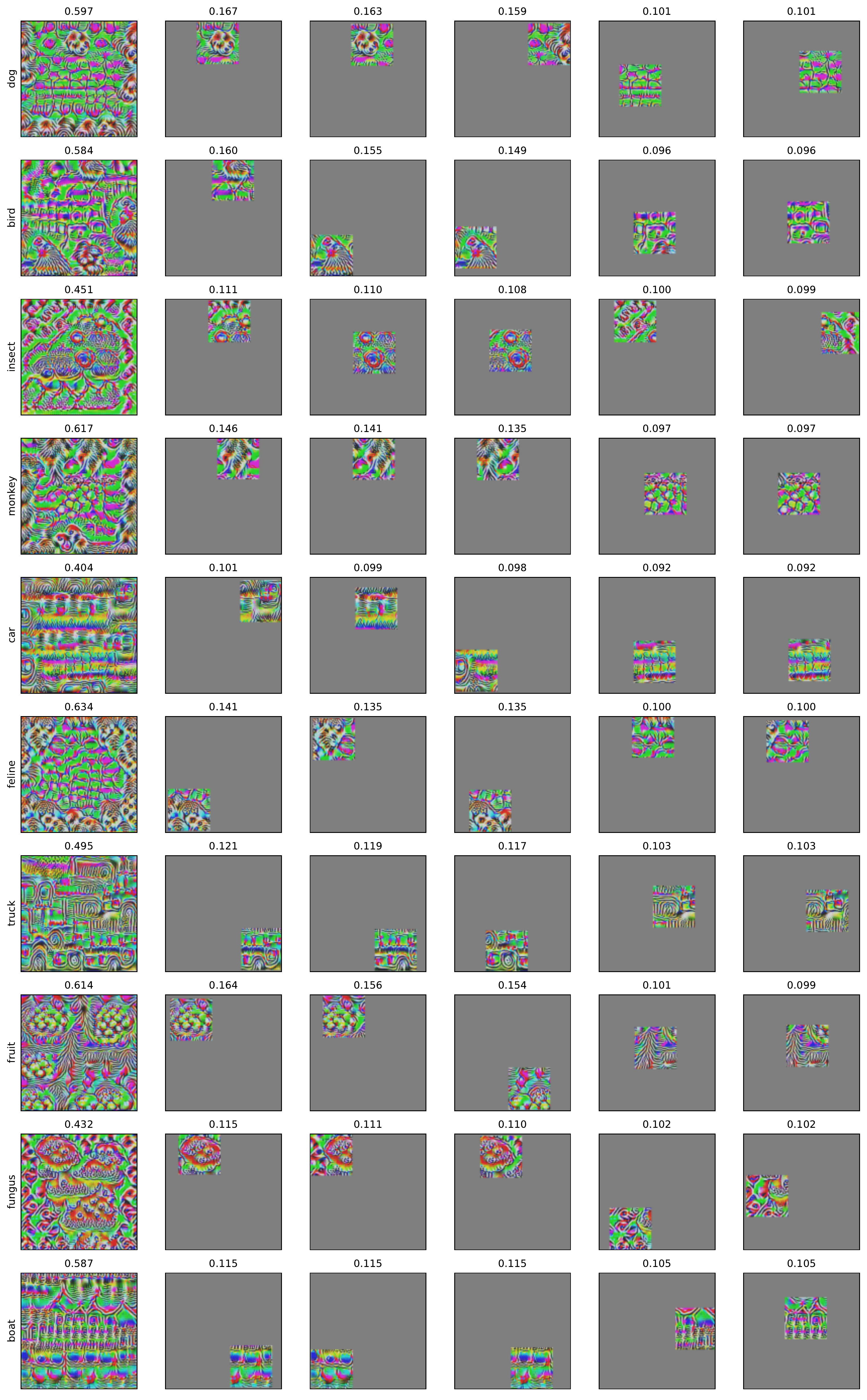}
  \caption{Locality analysis of $\linf$ universal perturbations on ImageNet-M10. Each row shows the perturbation for the particular class, followed by three patches with highest ASR and two patches with lowest ASR; ASR is written above each patch}
  \label{fig:full_locality_m10_linf}
\end{figure*}

We provide full results of randomized locality analysis on ImageNet-M10
in \Cref{fig:full_locality_m10_l2,fig:full_locality_m10_linf}.
We did not perform the analysis on CIFAR-10 perturbations, as the perturbations
are more ``global'' by nature, e.g. the semantic part seems to already occupy most of the perturbation.

\textbf{Exploring properties of $\ltwo$ vs $\linf$ perturbations.} For both $\ltwo$ and $\linf$ universal perturbations,
the patches that are most identifiable with their classes have much higher ASR.
A major difference is that for $\ltwo$ perturbations, the patches with low ASR have
much smaller norms, whereas for $\linf$ perturbations, the patches with low ASR have the similar norms
(measured in either $\ltwo$ or $\linf$).
This is expected, as for $\ltwo$ every local patch contributes to the total $\ltwo$ norm and it is unsurprising that optimization avoid putting weight on patches with little signal; in contrast,
for $\linf$ perturbations, different patches are independent in meeting the norm bound, so the perturbation can have patches with low signal that are similar in norm.
However, it is not clear whether these low signal patches found by optimization serve some purpose.
To test this, we experimented with zeroing-out these patches or replacing them with copies of patches with the highest ASR. Evaluation of the resulting perturbations show that the ASR is highest with these patches in place, and are in fact lowered when we replace them with patches with higher ASR.
This suggests that these low signal patches have a non-linear interaction with rest of the perturbation, and are providing a small boost in signal when combined with other patches.
Understanding this phenomena in more depth may shed more light on the differences between properties of $\linf$ and $\ltwo$ universal perturbations.

\section{Spatial Invariance}
\label{app:spatial}

\Cref{sec:spatial} shows a sample of translation invariance evaluation for perturbations on ImageNet-M10.
We include additional samples from evaluation using $\ltwo$ perturbations on CIFAR-10, to check that the phenomena
is general.

\begin{figure*}[!htbp]
  \centering
  \includegraphics[width=\linewidth]{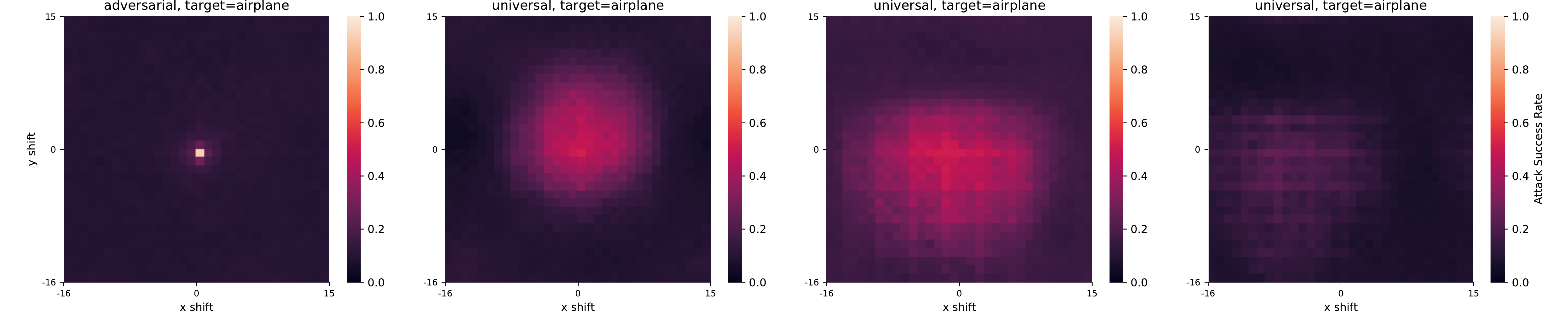}
  \caption{
    We evaluate attack success rates of translated adversarial and
    universal perturbations for different target classes over the test set on CIFAR-10.
    There is no subsampling of the grid here as the dimensions are much smaller.
    We used a subsampled grid of all the possible offsets with strides of four pixels.
    As before, the value at coordinate $(i, j)$ represents the average attack success rate when the
    perturbations are shifted by $i$ pixels along the $x$-axis and $j$ pixels
    along the $y$-axis, with wrap around to preserve information. Evaluation of adversarial perturbation
    is only shown for one target as it is redundant.
  }
  \label{fig:spatial_app}
\end{figure*}

\section{Perturbation Diversity}
\label{app:diversity}

\begin{figure*}[!htbp]
  \includegraphics[width=\linewidth]{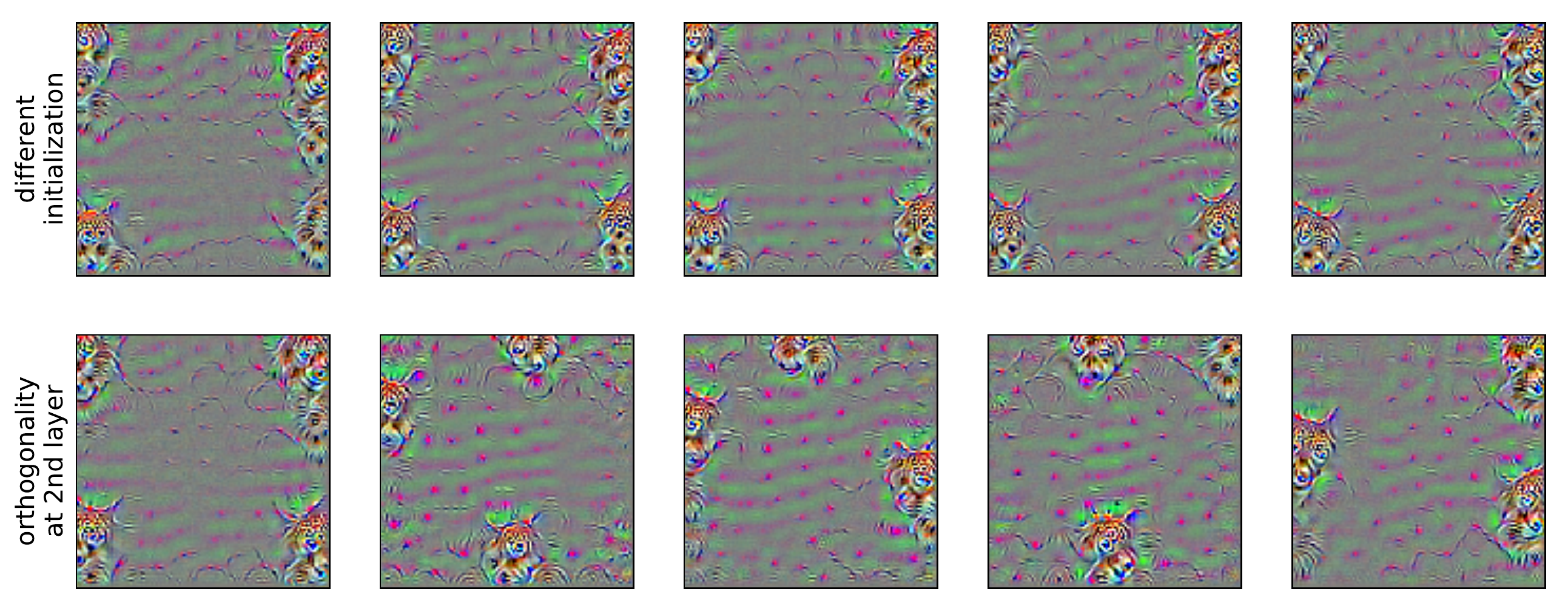}
  \caption{Visualization of $\ell_2$ universal perturbations for class feline, computed using different initializations (first row), and an additional penalty to encourage diversity (second row).}
  \label{fig:diversity}
\end{figure*}

Throughout our investigations, we observed that the computed universal perturbations
seemed to lack in diversity.
We see in Figure~\ref{fig:diversity} (first row) that generating
multiple different universal perturbations for a given target
leads to visually similar perturbations.\footnote{In the original work on universal perturbations, \cite{moosavi2017universal}
made the opposite interpretation that the perturbations generated independently are diverse due to their inner product being small. However, it is unclear whether inner product in the input space is the appropriate measure of their similarity. In fact, given the evidence of weak signal in universal perturbations (\Cref{sec:signal}), it appears more likely that different perturbations are highly similar in their (signal) content.}
Prior works study generating more diverse universal perturbations \citep{reddy2018nag}.

To increase diversity, we generated universal perturbations with an additional penalty to
encourage orthogonality of perturbed representations at different layers of the network.\footnote{For an extensive survey of similar and other regularization techniques for visualization, see \cite{olah2017feature}.} Even
with the added regularization, the resulting perturbations appear to be remain similar, but with changes in spatial arrangements of the semantic patterns (second row).

\begin{figure*}[!htbp]
  \includegraphics[width=\linewidth]{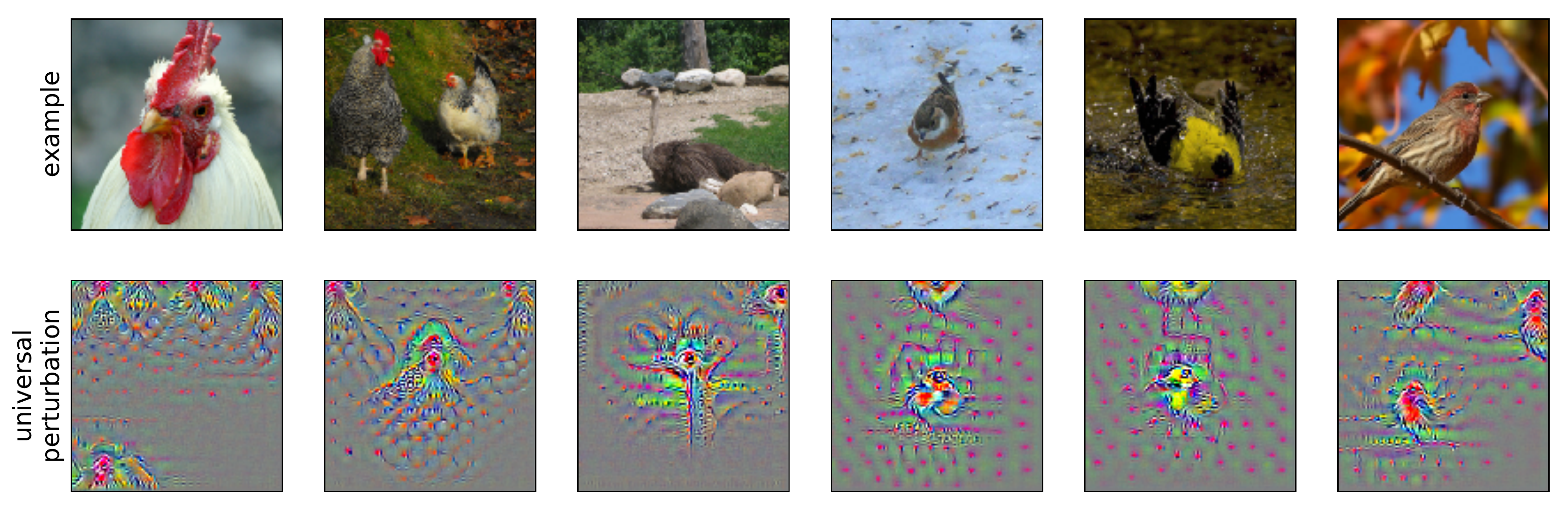}
  \caption{Visualization of $\ell_2$ universal perturbations for six individual sub-classes inside the \emph{bird} class. These were generated on a \emph{fine-grained} model trained on the subclass labels (60 in total). Observe the distinctive visual characteristics of each class, for instance the outline of a chicken (column 2), the neck of an ostrich (column 3), and the distinctive colors of different species (columns 3-5).}
  \label{fig:fg_ptbs}
\end{figure*}

One effective way we found to increase the diversity of these perturbations is to compute
universal perturbations for targeting specific \emph{sub-classes} of Mixed10.
To do so, we first train a fine-grained model using Mixed10's subclass labels, and generate
universal perturbations for each of the six sub-classes per class. In Figure~\ref{fig:fg_ptbs}, we show the universal perturbations targeting individual sub-classes of the bird
class. We observe distinct patterns characteristic of each
species of birds, an increase in diversity from before. However, generation of these perturbations requires a model trained on the fine-grained labels. An interesting question is whether one can recover such diversity using just the model trained on coarse-grained labels without injection of additional label information.

\end{document}